\documentclass{new_tlp}
\usepackage[utf8]{inputenc}
\usepackage{amssymb}
\usepackage{amsmath}
\usepackage{stmaryrd}

\usepackage[breaklinks=true]{hyperref}
\usepackage{breakcites}

\usepackage{pgf}

\usepackage{tikz}			
\usepackage{thmtools,thm-restate}	
\usetikzlibrary{shapes,arrows,automata}
\tikzstyle{IN}=[rectangle,draw,fill=green!25!,rounded corners]			
\tikzstyle{OUT}=[rectangle,draw,fill=red!40,rounded corners]			
\tikzstyle{UNDEC}=[rectangle,draw,fill=blue!60!green!20,rounded corners]	
\tikzstyle{MAIN}=[rectangle,draw,rounded corners]				
\tikzstyle{CONTROL}=[rectangle,rounded corners]				
\usetikzlibrary{arrows, shapes.geometric,fit, positioning}	

\newcommand{\set}[1]{\left\{#1\right\}}
\newcommand{\powerset}{\raisebox{.15\baselineskip}{\Large\ensuremath{\wp}}}

\newcommand{\comments}[1]{}

\newcommand{\unk}{\mathtt{unk}}

\newcommand{\naf}{\mathtt{not\ }}

\newcommand{\pr}{\mathit{par}}

\newcommand{\tr}{\mathbf{t}}
\newcommand{\fa}{\mathbf{f}}
\newcommand{\un}{\mathbf{u}}

\def\adf{\mathit{ADF}}
\def\nadf{\mathit{ADF}^+}
\def\hbp{\mathit{HB}_P}
\def\head{\mathit{head}}
\def\dom{\mathit{dom}}
\def\lp{\mathcal L_P}
\def\nlp{\textit{NLP}}
\def\setaf{\textit{SETAF}}
\newcommand{\Sup}{\mathtt{Sup}}
\newcommand{\Rules}{\mathtt{Rules}}
\newcommand{\Conc}{\mathtt{Conc}}

\def\aaf{\textit{AAF}}

\newtheorem{mdef}{Definition}

\newtheorem{example}[mdef]{Example}

\title[On the Equivalence Between $\adf$s and Logic Programs]
{On the Equivalence Between Abstract Dialectical Frameworks and Logic Programs}
\author[J. Alcântara and S. Sá and J. Acosta-Guadarrama]
{J. ALCÂNTARA and S. SÁ\\
Federal University of Ceará, Brazil\\
\email{jnando@lia.ufc.br and samy@ufc.br}
\and J. ACOSTA-GUADARRAMA\\
Autonomous University of Juarez, Mexico\\
\email{juan.acosta@uacj.mx}}

\begin{document}

\maketitle

\begin{abstract} 
Abstract Dialectical Frameworks ($\adf$s) are argumentation frameworks where each node is associated with an acceptance condition. This allows us to model different types of dependencies as supports and attacks. Previous studies provided a translation from Normal Logic Programs ($\nlp$s) to $\adf$s and proved the stable models semantics for a normal logic program has an equivalent semantics to that of the corresponding $\adf$. However, these studies failed in identifying a semantics for $\adf$s equivalent to a three-valued semantics (as partial stable models and well-founded models) for $\nlp$s. In this work, we focus on a fragment of $\adf$s, called Attacking Dialectical Frameworks ($\nadf$s), and provide a translation from $\nlp$s to $\nadf$s robust enough to guarantee the equivalence between partial stable models, well-founded models, regular models, stable models semantics for $\nlp$s and respectively complete models, grounded models, preferred models, stable models for $\adf$s. In addition, we define a new semantics for $\nadf$s, called $L$-stable, and show it is equivalent to the $L$-stable semantics for $\nlp$s. This paper is under consideration for acceptance in TPLP.
\end{abstract}

\section{Introduction}\label{s:introd}

Logic Programming and Formal Argumentation Theory are two different formalisms widely used for the representation of knowledge and reasoning. The connection between them is especially clear when comparing the semantics proposed to each formalism. The first questions were raised and answered in \cite{dung95acceptability}, the work that originally introduced Abstract Argumentation Frameworks (\aaf): it was shown how to translate a Normal Logic Program ($\nlp$) to an $\aaf$ and proved the stable models (resp. the well-founded model) of an $\nlp$ correspond to the stable extensions (resp. the grounded extension) of its corresponding $\aaf$. Other advances were made when \cite{wu2009complete} pointed the equivalence between the complete semantics for $\aaf$ and the partial stable semantics for $\nlp$s. Those semantics generalize many others, wielding a plethora of results gathered in \cite{caminada15equivalence}. 
One equivalence formerly expected to hold, however, could not be achieved: the correspondence between the semi-stable semantics for $\aaf$s \cite{caminada2006semi} and the $L$-stable semantics for $\nlp$s \cite{eiter97partial}. 

Despite their success, $\aaf$s are not immune to criticisms. A contentious issue refers to their alleged limited expressivity as they lack features which are common in almost every form of argumentation found in practice \cite{brewka2010abstract}. Indeed, in $\aaf$s the only interaction between atomic arguments is given by the attack relation. 

With such a motivation, in \cite{brewka2010abstract,brewka2013abstract} they defined Abstract Dialectical Frameworks ($\adf$s), a generalization of $\aaf$s, to express arbitrary relationships among arguments. In an $\adf$, besides the attack relation, arguments may support each other, or a group of arguments may jointly attack another while each argument in the group is not strong enough to do so. Such additional expressiveness arises by associating to each node (argument) its two-valued acceptance conditions which can get expressed as arbitrary propositional formulas. The intuition is that an argument is accepted if its associated acceptance condition is true.


A translation from $\nlp$s to $\adf$s is given in \cite{brewka2010abstract}, where they showed Stable Models Semantics for $\nlp$s has an equivalent semantics for $\adf$s.
However, they did not identify a semantics for $\adf$s equivalent to a 3-valued semantics (such as Partial Stable Models) for $\nlp$s \cite{brewka2010abstract,strass2013approximating}.

In this work, we will not only identify such semantics, but we will also ascertain only a fragment of $\adf$s, called Attacking Dialectical Frameworks ($\nadf$s), is needed. In fact, we will adapt the translation from $\nlp$s to Abstract Argumentation proposed in \cite{wu2009complete,caminada15equivalence} to provide a translation from $\nlp$s to $\nadf$s to account for various equivalences between their semantics. 
That includes to prove the equivalence between partial stable models, well-founded models, regular models, stable models semantics for $\nlp$s and respectively complete models, grounded models, preferred models, stable models for $\adf$s. 
Also, we define a new semantics for $\nadf$s, called $L$-stable (for least-stable), and show it is equivalent to the $L$-stable semantics for $\nlp$s \cite{eiter97partial}.
Hence, our results allow us to apply proof procedures and implementations for $\adf$s to $\nlp$s and vice-versa.


The paper proceeds as follows. 
Firstly we recall the basic definition of $\adf$s and $\nlp$s  as well as some of their well-established semantics. 
Next, we consider the Attacking Abstract Dialectical Frameworks  ($\nadf$s), a fragment of $\adf$s in which the unique relation involving arguments is the attack relation. 
In Section \ref{s:equiv}, we show a translation from $\nlp$s to $\nadf$s and prove the equivalence between partial stable models ($\nlp$s) and complete models ($\nadf$s), well-founded models ($\nlp$s) and grounded models ($\nadf$s), regular models ($\nlp$s) and preferred models ($\nadf$s), stable models ($\nlp$s) and stable models $\nadf$s, $L$-stable models ($\nlp$s) and $L$-stable models ($\nadf$s). In Section \ref{s:related}, we compare our results with previous attempts to translate $\nlp$s into $\adf$s and $\adf$s into $\nlp$s and we present a brief account on the main connections between $\nlp$s and Abstract Argumentation Frameworks \cite{dung95acceptability}/Assumption-Based Argumentation \cite{dung2009assumption} as well as a comparison between $\nadf$ and $\setaf$ \cite{nielsen2006generalization}, an extension of $\aaf$s to allow joint attacks on arguments.
Finally, we round off with a discussion of the obtained results and pointer for future works.

\section{Background}\label{s:background}

\subsection{Abstract Dialectical Frameworks} \label{ss:adf}
Abstract Dialectical Frameworks ($\adf$s) have been designed in \cite{brewka2010abstract,brewka2013abstract} to treat arguments (called statements there) as abstract and atomic entities. 
One can see it as a directed graph whose nodes represent statements, which can get accepted or not. 
Besides, the links between nodes represent dependencies: the status (accepted/not accepted) of a node $s$ only depends on the status of its parents ($\pr(s)$), i.e., the nodes with a direct link to $s$. We will restrict ourselves to finite $\adf$s: 

\begin{mdef}[Abstract Dialectical Frameworks \protect\cite{brewka2010abstract}]\label{d:adf}
An abstract dialectical framework is a tuple $D = (S, L, C)$ where 

\begin{itemize} 
\item $S$ is a finite set of statements (positions, nodes); 
\item $L \subseteq S \times S$ is a set of links, and $\forall s \in S$, $\pr(s) = \set{t \in S \mid (t,s) \in L}$; 
\item $C = \set{C_s \mid s \in S}$ is a set of total functions $C_s : 2^{\pr(s)} \to \set{\tr, \fa}$, one for each statement $s$. $C_s$ is called the acceptance condition of $s$. 
\end{itemize} 
\end{mdef} 

The function $C_s$ is intended to determine the acceptance status of a statement $s$, which only depends on the status of its parent nodes $\pr(s)$. Intuitively, $s$ will be accepted if there exists $R \subseteq \pr(s)$ such that $C_s(R) = \tr$, which means every statement in $R$ is accepted while each statement in $\pr(s) - R$ is not accepted. The acceptance conditions in $C$ of an $\adf$ $D = (S, L, C)$ can as well be represented in two alternative ways: 

\begin{itemize} 

\item Any function $C_s \in C$ can be represented by the set of subsets of $\pr(s)$ leading to acceptance, i.e., $C^\tr = \set{C^\tr_s \mid s \in S}$, where $C^\tr_s = \{R \subseteq \pr(s) \mid C_s(R) = \tr \}$. We will indicate this alternative by denoting an $\adf$ as $(S, L, C^\tr)$. 

\item Any function $C_s \in C$ can also be represented as a classical two-valued propositional formula $\varphi_s$ over the vocabulary $\pr(s)$ as follows: 
\begin{align}\label{eq:acceptance}
\varphi_s \equiv \bigvee_{R \in C^\tr_s} \left( \bigwedge_{a \in R} a \wedge \bigwedge_{b \in \pr(s) - R} \neg b \right).
\end{align}
If $C_s(\emptyset) = \tr$ and $\pr(s) = \emptyset$, we obtain $\varphi_s \equiv \tr$. If there is no $R \subset \pr(s)$ such that $C_s(R) = \tr$, then $\varphi_s \equiv \fa$. 
By $C^\varphi$ we mean the set $\set{\varphi_s \mid s \in S}$. We will indicate this alternative by denoting an $\adf$ as $(S, L, C^\varphi)$. We also emphasize any propositional formula $\varphi_s$ equivalent (in the classical two-valued sense) to the formula in Equation~(\ref{eq:acceptance}) can be employed to represent $C_s$. 
\end{itemize} 

When referring to an $\adf$ as $(S, L, C^\varphi)$, we will assume the acceptance formulas implicitly specify  the parents a node depends on. Then, the set $L$ of links between statements can be ignored, and the $\adf$ can be represented as $(S, C^\varphi)$, where $L$ gets recovered by $(t,s) \in L$ iff $t$ appears in $\varphi_s$. In order to define the different semantics for $\adf$s over the set of statements $S$, we will resort to the notion of (3-valued) interpretations:

\begin{mdef}[Interpretations and Models \protect\cite{brewka2010abstract}]
Let $D = (S, C^\varphi)$ be an $\adf$. 
A 3-valued interpretation (or simply interpretation) over $S$ is a mapping $v : S \to \set{\tr, \fa, \un}$ that assigns one of the truth values true ($\tr$), false ($\fa$) or unknown ($\un$),
to each statement. Interpretations will be extended to assign values to formulas over statements according to Kleene's strong 3-valued logic \cite{kleene1952introduction}:
negation switches $\tr$ and $\fa$, and leaves $\un$ unchanged; a conjunction is $\tr$ if both conjuncts are $\tr$, it is $\fa$ if some conjunct is $\fa$ and it is $\un$ otherwise; disjunction is dual. A 3-valued interpretation $v$ is a model of $D$ if for all $s \in S$ we have $v(s) \neq \un$ implies $v(s) = v(\varphi_s)$. 
\end{mdef}

Sometimes we will refer to an interpretation $v$ over $S$ as a set $V = \{s \mid s \in S \textit{ and } v(s) =$ $\tr \} \cup \set{\neg s \mid s \in S \textit{ and } v(s) = \fa}$. 
Obviously, if neither $s \in V$ nor $\neg s \in V$, then $v(s) = \un$. 

Furthermore, the three truth values are partially ordered by $\leq_i$ according
to their information content: $\un <_i \tr$ and $\un <_i \fa$ and
no other pair is in $<_i$. The pair $(\set{\tr, \fa, \un} , \leq_i )$ forms a complete meet-semilattice\footnote{A complete meet-semilattice is such that every non-empty finite subset has a greatest lower bound, the meet; and every nonempty
directed subset has a least upper bound. 
A subset is directed if any
two of its elements have an upper bound in the set.} with the meet operation $\sqcap$. This meet can be read
as consensus and assigns $\tr \sqcap \tr = \tr$, $\fa \sqcap \fa = \fa$, and returns $\un$ otherwise.

The information ordering $\leq_i$ extends as usual to interpretations $v_1$, $v_2$ over $S$ such that $v_1 \leq_i v_2$ iff $v_1(s) \leq_i v_2(s)$ for all $s \in S$. 
The set of all 3-valued interpretations over $S$ forms a complete meet-semilattice with respect to $\leq_i$. 
The consensus meet operation $\sqcap$ of this semilattice is given by $(v_1 \sqcap v_2)(s)$ = $v_1(s) \sqcap v_2(s)$ for all $s \in S$. 
The least element of this semilattice is the interpretation $v$ such that $v(s) = \un$ for each $s \in S$.

In \cite{brewka2013abstract}, the semantics for $\adf$s were defined via an operator $\Gamma_D$:

\begin{mdef}[$\Gamma_D$ Operator \protect\cite{brewka2013abstract}]\label{d:gamma}
Let $D = (S, L, C^\varphi)$ be an $\adf$ and $v$ be a 3-valued interpretation over $S$. We have
\[ \Gamma_D(v)(s) = \bigsqcap\set{w(\varphi_s) \mid w \in [v]_2}, \]
in which $[v]_2 = \set{w \mid v \leq_i w \textit{ and for each } s \in S, w(s) \in \set{\tr, \fa}}$.
\end{mdef}

Each element in $[v]_2$ is a 2-valued interpretation extending $v$. The elements of $[v]_2$ form an $\leq_i$-antichain with greatest lower bound $v = \bigsqcap[v]_2$. For each $s \in S$, $\Gamma_D$ returns the consensus truth value for $\varphi_s$, where the consensus takes into account all possible 2-valued interpretations $w$ extending $v$. If $v$ is 2-valued, we get $[v]_2 = \set{v}$. In this case, $\Gamma_D(v)(s) = v(\varphi_s)$ and $v$ is a 2-valued model for $D$ iff $\Gamma_D(v) = v$. As $[v]_2$ has only 2-valued interpretations, if $\varphi^1_s$ is equivalent to $\varphi^2_s$ in the classical two-valued sense, it is clear
\[ \bigsqcap\set{w(\varphi^1_s) \mid w \in [v]_2} = \bigsqcap\set{w(\varphi^2_s) \mid w \in [v]_2}. \]

That means when defining $\Gamma_D$ operator, it does not matter the acceptance formula we choose as far as it is equivalent in the classical 2-valued sense. In addition, $\Gamma_D$ operator can be employed to characterize also complete interpretations:

\begin{mdef}[Complete Interpretations \protect\cite{brewka2013abstract}]\label{d:admissible}
Let $D = (S, L, C^\varphi)$ be an $\adf$ and $v$ be a 3-valued interpretation over $S$. We state $v$ is a \emph{complete} interpretation of $D$ iff $v = \Gamma_D(v)$.
\comments{
\begin{itemize}
    \item $v$ is an \emph{admissible} interpretation of $D$ iff $v \leq_i \Gamma_D(v)$.
    \item $v$ is a \emph{complete} interpretation of $D$ iff $v = \Gamma_D(v)$.
\end{itemize}
}
\end{mdef}

As shown in \cite{brewka2010abstract}, $\Gamma_D$ operator is $\leq_i$-monotonic. 
Then a $\leq$-least fixpoint of $\Gamma_D$ is always guaranteed to exists for every $\adf$ $D$. 
Note complete interpretations of $D$ are also models of $D$.  
For this reason, they are also called complete models. The notion of reduct borrowed from logic programming \cite{gelfond1988stable} is reformulated to deal with $\adf$s:

\begin{mdef}[Reduct \protect\cite{brewka2013abstract}] Let $D = (S, L, C^\varphi)$ be an $\adf$ and $v$ be a 2-valued model of $D$. The reduct of $D$ with $v$ is given by the $\adf$, $D^v = (E_v, L^v, C^v)$, in which $E_v = \set{s \in S \mid v(s) = \tr}$, $L^v = L \cap (E_v \times E_v)$, and $C^v = \set{\varphi^v_s \mid s \in E_v \textit{ and } \varphi^v_s = \varphi_s[b \slash \fa : v(b) = \fa]}$; i.e., in each acceptance formula, $\varphi^v_s$, we replace in $\varphi_s$ every statement $b \in S$ by $\fa$ if $v(b) = \fa$.
\end{mdef}

We can now define some of the main semantics for an $\adf$ as follows:

\begin{mdef}[Semantics \protect\cite{brewka2013abstract}]\label{d:semantics}
Let $D = (S, L, C^\varphi)$ be an $\adf$, and $v$ a model of $D$. 
We state that
\begin{itemize}
    \item $v$ is a \emph{grounded model} of $D$ iff $v$ is the $\leq_i$-least complete model of $D$.
    \item $v$ is a \emph{preferred model} of $D$ iff $v$ is a $\leq_i$-maximal complete model of $D$.
    \item $v$ is a \emph{stable model} of $D$ iff $v$ is a 2-valued model of D such that $v$ is the grounded model of $D^v = (E_v, L^v, C^v)$.
\end{itemize}
 
\end{mdef}

We proceed by displaying an example to illustrate these semantics:

\begin{example}\label{ex:adf}
Consider the $\adf$, $D = (S, C^\varphi)$, given by
$a [\neg b] \qquad b[\neg a] \qquad c [\neg b \wedge e] \qquad d [\neg c] \qquad e [\neg d]$,
where $S= \set{a, b, c, d, e}$, and the acceptance formula of each $s \in S$ is written in square brackets on the right of $s$. As for the semantics for $D$, we have a) $\set{a, \neg b}$, $\set{b, d, \neg a, \neg c, \neg e}$ and $\emptyset$ are its complete models; b) $\emptyset$ is its grounded model; c) $\set{a, \neg b}$, $\set{b, d, \neg a, \neg c, \neg e}$ are its preferred models; d) $\set{b, d, \neg a, \neg c, \neg e}$ is its unique stable model.
\end{example}

Notice some $\adf$s have no stable models. For instance, in an $\adf$ whose unique statement is $a [\neg a]$, there is no stable model. Furthermore, an $\adf$ can have more than one stable model as the $\adf$ represented by $a[\neg b]$ and $b [\neg a]$,
in which $\set{a, \neg b}$ and $\set{b, \neg a}$ are the stable models of $D$. In contrast, the grounded model is unique for each $\adf$ (see \cite{brewka2010abstract,brewka2013abstract}).

\subsection{Normal Logic Programs}\label{ss:lp}

Now we will focus on propositional normal logic programs.
We assume the reader is familiar with the Stable Model Semantics~\cite{gelfond1988stable}.
\begin{mdef} \label{def-lp}
A Normal Logic Program ($\nlp$), $P$, is a set of rules of the form $a \leftarrow a_1, \ldots, a_m,\naf b_1,$ $\ldots, \naf b_n$ ($m, n \in \mathbb{N}$), where $a$, $a_i$ ($1 \leq i \leq m$) and $b_j$ ($1 \leq j \leq n$) are atoms; $\mathit{not}$ represents default negation, and $\naf b_j$ is a default literal. We say $a$ is the head of the rule, and $a_1, \ldots, a_m, \naf b_1, \ldots, \naf b_n$ is its body. The Herbrand Base of $P$ is the set $\hbp$ of all atoms occurring in $P$.
\end{mdef}

A wide range of logic programming semantics can be defined based on the 3-valued interpretations (for short, interpretations) of programs:

\begin{mdef}[Interpretation and Models \protect\cite{przymusinski90well-founded}]\label{d:interpretations} A 3-valued interpretation, $I$, of an $\nlp$, $P$, is a total function $I : \hbp \to \set{\tr, \fa, \un}$. We say $I$ is a model of $P$ iff for each rule $a \leftarrow a_1, \ldots, a_m,$ $\naf b_1, \ldots, $ $\naf b_n \in P$, $\mathit{min}\set{I(a_1), \ldots, I(a_m), \neg I(b_1), \ldots, \neg I(b_n)} \leq_t I(a)$, where $\neg \tr = \fa$, $\neg \fa = \tr$ and $\neg \un = \un$.

When convenient, we will refer to an interpretation $I$ of $P$ as a set $\mathcal I = \left\{a \mid \hbp  \textit{ and } \right.$ $\left. I(a) = \tr \right\} \cup \set{\neg a \mid a \in \hbp \textit{ and } I(a) = \fa}$. If neither $a \in \mathcal I$ nor $\neg a \in \mathcal I$, then $I(a) = \un$. 
\end{mdef}

Besides the information ordering $\leq_i$, it is worth mentioning here the truth ordering $\leq_t$ given by $\fa <_t \un <_t \tr$. The truth ordering $\leq_t$ extends as usual to interpretations $I_1$, $I_2$ over $\hbp$ such that $I_1 \leq_t I_2$ iff $I_1(a) \leq_t I_2(a)$ for all $a \in \hbp$. We also emphasize the notions of model of a logic program and model of an $\adf$ follow distinct motivations: the models of a logic program are settled on $\leq_t$ whereas the models of an $\adf$ are settled on $\leq_i$. In order to avoid confusions, we will let it explicit when referring to one of them.

Now we will consider the main semantics for $\nlp$s. Let $I$ be a 3-valued interpretation of a program $P$; take $P/I$ to be the program built by the execution of the following steps:

\begin{enumerate}
  \item Remove any $a \leftarrow a_1, \ldots, a_m,$ $\naf b_1, \ldots, \naf b_n \in P$ such that $I(b_i) = \tr$ for some $i$ ($1 \leq i \leq n$);
  \item Afterwards, remove any occurrence of $\naf b_i$ from $P$ such that $I(b_i) = \fa$.
  \item Then, replace any occurrence of $\naf b_i$ left by a special atom $\mathbf{u}$ ($\mathbf{u} \not\in \hbp$).
\end{enumerate}

Note $\mathbf{u}$ is assumed to be unknown in each interpretation of $P$. As shown in \cite{przymusinski90well-founded}, $P/I$ has a unique $\leq_t$-least 3-valued model, obtained by the $\Psi$ operator:

\begin{mdef}[$\Psi_{\frac{P}{I}}$ Operator \protect\cite{przymusinski90well-founded}] Let $P$ be an $\nlp$, $I$ and $J$ be interpretations of $P$ and $a \in \hbp$ an atom in $P$. Define $\Psi_{\frac{P}{I}}(J)$ to be the interpretation given by
\begin{itemize}
  \item $\Psi_{\frac{P}{I}}(J)(a) = \tr$ if $a \leftarrow a_1, \ldots, a_m \in P/I$ and for all $i$, $1 \leq i \leq m$, $J(a_i) = \tr$;
  \item $\Psi_{\frac{P}{I}}(J)(a) = \fa$ if for every $a \leftarrow a_1, \ldots, a_m \in P/I$, there exists $i$, $1 \leq i \leq m$, such that $J(a_i) = \fa$;
  \item $\Psi_{\frac{P}{I}}(J)(a) = \un$ otherwise.
\end{itemize}

\end{mdef}

Indeed, the $\leq_t$-least model of $\frac{P}{I}$, denoted by $\Omega_P(I)$, is given by the least fixed point of $\Psi_{\frac{P}{I}}$ iteratively obtained as follows for finite logic programs:
\begin{align*}
    \Psi^{\uparrow\ 0}_{\frac{P}{I}} = & \bot\\ 
    \Psi^{\uparrow\ i + 1}_{\frac{P}{I}} = & \Psi_{\frac{P}{I}}(\Psi^{\uparrow\ i}_{\frac{P}{I}})
\end{align*}

\noindent in which $\bot$ is an interpretation such that for each $a \in \hbp$, $\bot(a) = \fa$. According to \cite{przymusinski90well-founded}, there exists $n \in  \mathbb{N}$ such that $\Omega_P(I) = \Psi^{\uparrow\ n + 1}_{\frac{P}{I}} = \Psi^{\uparrow\ n}_{\frac{P}{I}}$. We now specify the logic programming semantics to be examined in this paper.

\begin{mdef} \label{def-lpsemantics} Let $P$ be an $\nlp$ and $I$ be an interpretation:

\begin{itemize}
  \item $I$ is a partial stable model (\textit{PSM}) of $P$ iff $I = \Omega_P(I)$ \cite{przymusinski90well-founded}.
  \item $I$ is a well-founded model of $P$ iff $I$ is the $\leq_i$-least \textit{PSM} of $P$ \cite{przymusinski90well-founded}.
  \item $I$ is a regular model of $P$ iff $I$ is a $\leq_i$-maximal \textit{PSM} of $P$  \cite{eiter97partial}.
  \item $I$ is a stable model of $P$ iff $I$ is a \textit{PSM} of $P$ where for each $a \in \hbp$, $I(a) \in \set{\tr, \fa}$ \cite{przymusinski90well-founded}.
  \item $I$ is an $L$-stable model of $P$ iff $I$ is a \textit{PSM} of $P$ with minimal $\set{a \in \hbp \mid I(a) = \un}$ (w.r.t. set inclusion) among all partial stable models of $P$ \cite{eiter97partial}.
\end{itemize}

\end{mdef}
\begin{example}\label{ex:nlp}
Consider the $\nlp$ $P$:
$$
\begin{array}{llllll}
	b \leftarrow c, \naf a\quad &  a \leftarrow \naf b\quad & c \leftarrow d\quad &	p \leftarrow c, d, \naf p\quad & p \leftarrow \naf a\quad & d \leftarrow 
\end{array}
$$
Concerning the semantics of $P$, we have a) Partial stable models: $\set{c, d}$, $\set{b, c, d, p, \neg a}$ and $\set{a, c, d, \neg b}$; b) Well-founded model: $\set{c, d}$; c) Regular models: $\set{b, c, d, p, \neg a}$ and $\set{a, c, d, \neg b}$; d) Stable model and $L$-Stable model: $\set{b, c, d, p, \neg a}$.
\end{example}

In the next section, we will focus on a fragment of $\adf$s, dubbed Attacking Abstract Dialectical Frameworks ($\nadf$s), and in the sequel we will show that $\nadf$s are enough to capture any semantics based on partial stable models as those above mentioned.

\section{Attacking Abstract Dialectical Frameworks}\label{s:attacking}

Now we consider the Attacking Abstract Dialectical Frameworks  ($\nadf$s), a fragment of $\adf$s in which the unique relation involving statements is the attack relation. We may note parenthetically some definitions related to $\adf$s become simpler when restricted to $\nadf$s. We proceed by recalling the notions of supporting and attacking links:

\begin{mdef}[Supporting and Attacking Links \protect\cite{brewka2010abstract}]\label{d:supattlinks}
Let $D = (S, L, C)$ be an $\adf$. A link $(r, s) \in L$ is
\begin{description}
    \item \emph{supporting in $D$} iff for no $R \subseteq \pr(s)$ we have $C_s(R) = \tr$ and $C_s(R \cup \set{r}) = \fa$.
    \item \emph{attacking  in $D$} iff for no $R \subseteq \pr(s)$ we have $C_s(R) = \fa$ and $C_s(R \cup \set{r}) = \tr$.
\end{description}

\end{mdef}

Formally, a link $(r,s)$ is \emph{redundant} if it is both attacking and supporting. Redundant links can be deleted from an $\adf$ as they mean no real dependencies \cite{brewka2010abstract}. 
Again in \cite{brewka2010abstract}, the authors introduced the Bipolar Abstract Dialectical Frameworks ($\mathit{BADF}$), a subclass of $\adf$s in which every link is either supporting or attacking. Now we regard a subclass of $\mathit{BADF}s$ in which only attacking links are admitted:

\begin{mdef}[$\nadf$]\label{def:nadf} An Attacking Abstract Dialectical Framework, denoted by $\nadf$, is an $\adf$ $(S, L, C)$ such that every $(r,s) \in L$ is an attacking link. This means that for every $s \in S$, if $C_s(M) = \tr$, then for every $M' \subseteq M$, we have $C_s(M') = \tr$. 
\end{mdef}

In an $\nadf$ $(S, L, C)$, for each $s \in S$, its acceptance formula $\varphi_s$ can be simplified as follows:

\begin{restatable}{theo}{nnadf}\label{t:nadf} Let $D = (S, L, C^\tr)$ be an $\nadf$ and, for every $s \in S$, we define $C^\mathit{max}_s = \left\{R \in C^\tr_s \mid \textit{there is} \right.$ $\left. \textit{no } R' \in C^\tr_s \textit{ such that } R \subset R' \right\}$. 
Then, for every $s \in S$,
\[ \varphi_s \equiv \bigvee_{R \in C^\mathit{max}_s}  \bigwedge_{b \in \pr(s) - R} \neg b.\]
\end{restatable}

Hence, in $\nadf$s, every acceptance formula corresponds to a propositional formula in the disjunctive normal form, where each disjunct is a conjunction of negative atoms. 
Notice replacing an acceptance formula by a two-valued equivalent one does not change the complete semantics, and we are not interested in the three-valued models of the $\nadf$.
The importance of these formulas will be evident below. 
Before, however, note $\nadf$ does not prohibit redundant links. 
For instance, consider the $\adf$ $D = (S, L, C)$, in which $S = \set{a, b, c}$, $L = \set{(b,a), (c, a)}$ and $C^\tr_a = \set{\set{b}, \emptyset}$ and $C^\tr_b = C^\tr_c = \set{\emptyset}$. We know $D$ is an $\nadf$ as both $(b,a)$ and $(c,a)$ are attacking links. In addition, $(b,a)$ is a redundant link as it is also supporting. Redundant links can be easily identified in $\nadf$s:

\begin{restatable}{theo}{redundant}\label{t:redundant} Let $D = (S, L, C^\tr)$ be an $\nadf$.  A link $(r,s) \in L$ is redundant iff $r \in R$ for every $R \in C^\mathit{max}_s$.
\end{restatable}

A straightforward consequence from Theorem \ref{t:redundant} is that in $\nadf$s, every acceptance formula $\varphi_s$ in the disjunctive normal form as in Theorem \ref{t:nadf}, where each disjunct is a conjunction of negative atoms, disregards redundant links:

\begin{restatable}{cor}{redundantII}\label{c:redundant2} Let $D = (S, L, C^\tr)$ be an $\nadf$. For each $s \in S$, if $\varphi_s$ is $\displaystyle \bigvee_{R \in C^\mathit{max}_s}  \bigwedge_{b \in \pr(s) - R} \neg b$ 
and $L' = \set{(r,s) \mid \neg r \textit{ appears in } \varphi_s}$, then $L'$ has no redundant link. 
\end{restatable}

\begin{example}
Let us recall the $\nadf$ $D = (S, L, C)$ above in which $S = \set{a, b, c}$, $L = \set{(b,a), (c, a)}$ and $C^\tr_a = \set{\set{b}, \emptyset}$ and $C^\tr_b = C^\tr_c = \set{\emptyset}$. 
With the general representation for $\varphi_s$ in $\adf$s, in which for every $s \in S$, $\displaystyle \varphi_s \equiv \bigvee_{R \in C^\tr_s} \left( \bigwedge_{a \in R} a \wedge \bigwedge_{b \in \pr(s) - R} \neg b \right)$, we get
$a [(b \wedge \neg c) \vee (\neg b \wedge \neg c)] \qquad b[\tr] \qquad c [\tr]$. 
With the simpler representation for acceptance formulas given by Theorem \ref{t:nadf}, in which $\displaystyle \varphi_s \equiv \bigvee_{R \in C^\mathit{max}_s}  \bigwedge_{b \in \pr(s) - R} \neg b$, we get $a [\neg c] \qquad b[\tr] \qquad c [\tr]$. As expected, the redundant link $(b,a)$ is not taken into account to define $\varphi_a$ as $\neg c$.
\end{example}

Alternatively, redundant links in $\nadf$s have the following property:

\begin{restatable}{theo}{redundantIII}\label{t:redundant3} Let $D = (S, L, C^\tr)$ be an $\nadf$, $s \in S$; $r \in \pr(s)$ and $C^\tr_s(r) = \set{R \in C^\tr_s \mid r \in R}$.  A link $(r,s) \in L$ is redundant iff $ | C^\tr_s(r) | = \dfrac{|C^\tr_s|}{2}$.
\end{restatable}

Thus, identifying redundant links in an $\nadf$ has a sub-quadratic time complexity on $|C^\tr_s|$:

\begin{restatable}{cor}{linear}\label{c:linear} Let $D = (S, L, C^\tr)$ be an $\nadf$. Deciding if a link $(r,s) \in L$ is redundant can be solved in sub-quadratic time on $|C^\tr_s|$.
\end{restatable}

In contrast, identifying redundant links in $\adf$s is coNP-hard \cite{ellmauthaler2012}.

In Subsection \ref{ss:adf}, $\Gamma_D$ operator is employed to define the semantics for $\adf$. 
When restricted to $\nadf$s, it assumes a simpler version:

\begin{restatable}{theo}{gammaomega}\label{t:gammaomega}
 Let $D = (S, L, C^\varphi)$ be an $\nadf$, $v$ be a 3-valued interpretation over $S$, and for each $s \in S$,  $\varphi_s$ is the formula $\displaystyle \bigvee_{R \in C^\mathit{max}_s}  \bigwedge_{b \in \pr(s) - R} \neg b$ depicted in Theorem \ref{t:nadf}. It holds for every $s \in S$, $\Gamma_D(v)(s) = v(\varphi_s)$.
\end{restatable}

Besides being noticeably simpler when restricted to $\nadf$, this new characterization of $\Gamma_D$ might mean lower complexity of reasoning. In \cite{brewka2013abstract}, the problem of verifying whether a given interpretation is complete is proved to be DP-complete. In our case, owing to our definition of $\Gamma_D$, this problem can get solved by assigning values to formulas over statements according to Kleene's strong 3-valued logic. 
This evaluation procedure is similar to (and has the same complexity as) that for Boolean formulas, which takes polynomial time \cite{buss1987boolean}. 
We run this procedure for each statement in a given $\adf$. 
Then, the overall algorithm runs in polynomial time. 
It is a promising result as the complexity of many reasoning tasks on $\nadf$s may likely have the same complexity as standard Dung's $\aaf$s \cite{dung95acceptability}. 
A consequence from Theorem~\ref{t:gammaomega} is the stable models of an $\nadf$ $D$ can get characterized as the two-valued complete models of $D$:

\begin{restatable}{theo}{stable}\label{t:stable}
 Let $D = (S, L, C^\varphi)$ be an $\nadf$.
 Then $v$ is a stable model of $D$ iff $v$ is a 2-valued complete model of $D$.
\end{restatable}

The main objective of this work is to show each semantics for $\nlp$s presented in Subsection~\ref{ss:lp} has an equivalent one for $\nadf$. Then we need to define a new semantics for $\nadf$, which will be proved in the next section to be equivalent to the $L$-stable models semantics for $\nlp$s:

\begin{mdef}[L-stable]
 Let $D = (S, L, C^\varphi)$ be an $\nadf$, and $v$ be a 3-valued interpretation of $D$. We say $v$ is an $L$-stable model of $D$ iff $v$ is a complete model with minimal $\unk(v) = \set{s \in S \mid v(s) = \un}$ (w.r.t. set inclusion) among all complete models of $D$. 
\end{mdef}

Note $L$-stable models semantics is defined for every $\nadf$ and the $L$-stable models of an $\nadf$ $D$ will coincide with its stable models whenever $D$ has at least one stable model. 
Indeed we can see a stable model $v$ as an $L$-stable model in which $\unk(v) = \emptyset$. 


\begin{example}\label{ex:adf2}
Consider the $\nadf$ $D = (S, C^\varphi)$ given by
\[ a [\neg b] \qquad b[\neg a] \qquad c [(\neg c \wedge \neg a) \vee (\neg c \wedge \neg d)] \qquad d[\neg d] \qquad e[\neg e \wedge \neg b],  \]
where $S= \set{a, b, c, d, e}$, and the acceptance formula of each statement $s \in S$ is written in square brackets on the right of $s$. As for the semantics of $D$, a) $\set{a, \neg b}$, $\set{b, \neg a, \neg e}$ and $\emptyset$ are its complete models; b) $\emptyset$ is its grounded model; c) $\set{a, \neg b}$ and $\set{b, \neg a, \neg e}$ are its preferred models;  d) $D$ has no stable model; e) $\set{b, \neg a, \neg e}$ is its unique $L$-stable model.
\end{example}

Thus none of these semantics for $\nadf$ are equivalent to each other. However, in the sequel, we will show some equivalences between $\nlp$s semantics and $\nadf$ semantics.

\comments{
\begin{restatable}{theo}{completenadf}\label{t:completenadf} Let $D = (S, L, C^\tr)$ be an $\nadf$. A three-valued interpretation $v$ is a complete model of $D$ iff for each $s \in S$ it holds 

\begin{itemize}
    \item $v(s) = \tr$ iff there exists $K \subseteq C^\tr_s$ such that for each $b \in \pr(s) - K$, it holds $v(b) = \fa$. 
    \item $v(s) = \fa$ iff for each $K \subseteq C^\tr_s$,  there exists $b \in \pr(s) - K$ such that $v(b) = \tr$. 
    \item $v(s) = \un$ iff for each $K \subseteq C^\tr_s$, there exists $b \in \pr(s) - K$ such that $v(b) \neq \fa$, and there exists $K \subseteq C^\tr_s$ such that for every $b \in \pr(s) - K$, it holds $v(b) \neq \tr$.
\end{itemize}
\end{restatable}
}

\section{Equivalence Between ADF and Logic Programs}\label{s:equiv}

We will show one particular translation from $\nlp$ to $\nadf$ is able to account for a whole range of equivalences between their semantics. This includes to prove the equivalence between $\nlp$ partial stable models and $\nadf$ complete models, $\nlp$ well-founded models and $\nadf$ grounded models, $\nlp$ regular models and $\nadf$ preferred models, $\nlp$ stable models and $\nadf$ stable models, $\nlp$ $L$-stable models and $\nadf$ $L$-stable models. Our treatment is based on a translation from $\nlp$ to Abstract Argumentation proposed in \cite{wu2009complete,caminada15equivalence}, where each $\nlp$ rule is directly translated into an argument. In contradistinction, we will adapt it to deal with $\adf$ by translating each rule into a substatement, and then, substatements corresponding to rules with the same head are gathered to constitute a unique statement. Taking a particular $\nlp$ $P$, one can start to construct substatements recursively as follows:

\begin{mdef}[Substatement]\label{d:substatement}
Let $P$ be an $\nlp$.
\begin{itemize}
  \item If $a$ is a rule (fact) in $P$, then it is also a substatement (say $r$) in $P$ with $\Conc_P(r) = a$, $\Rules_P(r) = \{ a \}$ and $\Sup_P(r) = \{ \}$. 
  \item If $a \leftarrow \naf b_1, \ldots, \naf b_n$ is a rule in $P$, then it is also a substatement (say $r$) in $P$ with $\Conc_P(r) = a$, $\Rules_P(r) = \{ a \leftarrow \naf b_1, \ldots, \naf b_n \}$ and $\Sup_P(r) = \{ \neg b_1, \ldots, \neg b_n \}$.
  \item If $a \leftarrow a_1,\ldots,a_m, \naf b_1,\ldots,\naf b_n$ 
	is a rule in $P$ and for each $a_i$ ($1 \leq i \leq m$) 
	there exists a substatement $r_i$ in $P$  with $\Conc_P(r_i) = a_i$
	and $a \leftarrow a_1,\ldots,a_m, \naf b_1,\ldots,$ $\naf b_n$ is not contained in $\Rules_P(r_i)$,	then $a \leftarrow r_1,\ldots,r_m,\naf b_1,\ldots,\naf b_n$ is a substatement (say $r$) in $P$ with $\Conc_P(r) = a$, $\Rules_P(r) = \Rules_P(r_1) \cup \ldots \cup \Rules_P(r_m)\ \cup$
	$\{ a \leftarrow a_1,\ldots,a_m, \naf b_1,\ldots,\naf b_n \}$ and
	$\Sup_P(r) = \Sup_P(r_1)$ $\cup \ldots \cup$ $\Sup_P(r_n)$ $\cup$ $\{ \neg b_1, \ldots, \neg b_n \}$.
	\item Nothing more is a substatement in $P$.
\end{itemize}

\end{mdef}

For a substatement $r$ in $P$, $\Sup_P(r)$ is referred to as the \emph{support} of $r$ in $P$. Besides, for each substatement $r$ in $P$, we can also define $\Sup_P(r)$ iteratively as follows:
\begin{align*}
    \Sup^{\uparrow\ 0}_P(r)  = &\ \emptyset \\
    \Sup^{\uparrow\ i + 1}_P(r) = & \set{\neg b_1, \ldots, \neg b_n } \cup \Sup^{\uparrow\ i}_P(r_1) \cup \ldots \cup \Sup^{\uparrow\ i}_P(r_m) 
\end{align*}
such that $r$ is a substatement $a \leftarrow r_1,\ldots,r_m, \naf b_1,\ldots,\naf b_n$ in $P$, $a \leftarrow a_1,\ldots,a_m,$ $\naf b_1,\ldots,\naf b_n \in \Rules_P(r)$ and $\forall a_i$ ($1 \leq i \leq m$) there exists a substatement $r_i$ in $P$  with $\Conc_P(r_i) = a_i$. Note for each substatement $r$ in $P$, $\exists k \in \mathbb{N}$ such that $\Sup_P(r) = \Sup^{\uparrow\ k}_P(r)$. This notion of support is generalized to obtain the support of an atom in $P$:


\begin{mdef}[Support]\label{d:support}
Let $P$ be an $\nlp$ over a set $A$ of atoms. For each $a \in A$, we define the support of $a$ in $P$ as $\Sup_P(a)\hspace{-.2em} =\hspace{-.2em} \left\{\ \Sup_P(r)\hspace{-.2em}\ \mid\hspace{-.2em}\ r \textit{ is a substatement in }\hspace{-.1em} P\hspace{-.075em} \textit{ such that } \Conc_P(r) \right.$ $\left. = a\ \right\}$.
\end{mdef}

\begin{example}\label{ex:statements}
Consider the normal logic program $P$ from Example \ref{ex:nlp}:
$$
\begin{array}{llllll}
	b \leftarrow c, \naf a	\quad & a \leftarrow \naf b \quad & c \leftarrow d \quad p \leftarrow c, d, \naf p \quad &  p \leftarrow \naf a\quad & d \leftarrow
\end{array}
$$

We can obtain the following substatements:
\[
\begin{array}{llllll}
r_1:	&	d \leftarrow  & r_3:	&	p \leftarrow r_2, r_1, \naf p \qquad & r_5:	&	p \leftarrow \naf a \\
r_2:	&	c \leftarrow r_1 \qquad & r_4:	&	a \leftarrow \naf b &  r_6:	&	b \leftarrow r_2, \naf a.
\end{array}
\]
Thus
\[
\begin{array}{lll}
    \Sup_P(r_1) = \set{\ } \qquad  & \Sup_P(r_3) = \set{\neg p}\qquad &  \Sup_P(r_5) = \set{\neg a}\\
    \Sup_P(r_2) = \set{\ } & \Sup_P(r_4) = \set{ \neg b } &  \Sup_P(r_6)= \set{\neg a}.
\end{array}
\]
and
\[
\begin{array}{lll}
    \Sup_P(a) = \set{ \set{\neg b} } & \Sup_P(b) = \set{  \set{\neg a} }\qquad & \Sup_P(p) = \set{ \set{\neg p}, \set{\neg a} }\\
    \Sup_P(c) = \set{ \emptyset } & \Sup_P(d) = \set{\emptyset }. & \ \\
\end{array}
\]

\end{example}

After that, we can construct the corresponding $\adf$ as follows:

\begin{mdef}\label{d:xip}
Let $P$ be an $\nlp$ over a set $A$ of atoms. Define an $\adf$ $\Xi(P) = (A, L, C^\tr)$, in which

\begin{itemize}
    \item $L = \set{(b,a) \mid B \in \Sup_P(a) \textit{ and } \neg b \in B}$;
   \item For each $a \in A$, $C^\tr_a = \Big\{ B' \subseteq \set{b \in \pr(a) \mid \neg b \not\in B } \Big| B \in \Sup_P(a) \Big\}$.
\end{itemize}
\end{mdef}

We can prove the resulting $\adf$ $\Xi(P)$ is indeed an $\nadf$:

\begin{restatable}{prop}{xipnadf}\label{p:xipnadf}
Let $P$ be an $\nlp$.  The corresponding $\Xi(P)$ is an $\nadf$. 
\end{restatable}

Hence, the acceptance condition for each statement in $\Xi(P)$ can be retrieved as follows: 


\begin{restatable}{prop}{xipacceptance}\label{p:xipacceptance}
Let $P$ be an $\nlp$ and $\Xi(P) = (A, L, C^\tr)$ the corresponding $\nadf$. The acceptance condition $\varphi_a$ for each $a \in A$ is given by 
\[ \varphi_a \equiv \bigvee_{B \in \Sup_P(a)} \left( \bigwedge_{\neg b \in B} \neg b  \right). \]
In particular, if $\Sup_P(a) = \set{ \emptyset }$, then $\varphi_a \equiv \tr$ and if $\Sup_P(a) = \emptyset$, then $\varphi_a \equiv \fa$.
\end{restatable}

\begin{example}\label{ex:xip}
Recalling the $\nlp$ $P$ in Example \ref{ex:statements}, we obtain $\nadf$ $\Xi(P) = (A, L, C^\tr)$, in which $A = \set{a, b, c, d, p}$; $L = \set{(b,a), (a,b), (p,p), (a, p)}$;  $C^\tr_a = C^\tr_b = C^\tr_c = C^\tr_d = \set{\emptyset}$ and $C^\tr_p = \set{ \set{a}, \set{p}, \emptyset}$. The acceptance condition for each statement in $\Xi(P)$ is given below:
\[ a [\neg b] \qquad b[\neg a] \qquad c [\tr] \qquad d [\tr] \qquad p [\neg p \vee \neg a]. \]
Concerning the semantics of $\Xi(P)$, we have 

\begin{itemize}
    \item Complete models: $\set{c, d}$, $\set{b, c, d, p, \neg a}$ and $\set{a, c, d, \neg b}$;
    \item Grounded model: $\set{c, d}$;
    \item Preferred models: $\set{b, c, d, p, \neg a}$ and $\set{a, c, d, \neg b}$;
    \item Stable model and $L$-stable model: $\set{b, c, d, p, \neg a}$.
\end{itemize}
\end{example}

Now we can prove one of the main results of this paper: Partial Stable Models are equivalent to Complete Models. 

\begin{restatable}{theo}{pstable}\label{t:pstable}
Let $P$ be an $\nlp$ and $\Xi(P)$ be the corresponding $\nadf$. 
$v$ is a partial stable model of $P$ iff $v$ is a complete model of $\Xi(P)$.
\end{restatable}

With this equivalence showed in Theorem \ref{t:pstable}, the following results are immediate:

\begin{restatable}{theo}{equivalence}\label{t:equivalence}
Let $P$ be an $\nlp$ and $\Xi(P) = (A, L, C^\tr)$ the corresponding $\nadf$. We have 

\begin{itemize}
    \item $v$ is a well-founded model of $P$ iff $v$ is a grounded model of $\Xi(P)$.
    \item $v$ is a regular model of $P$ iff $v$ is a preferred model of $\Xi(P)$.
    \item $v$ is a stable model of $P$ iff $v$ is a stable model of $\Xi(P)$.
    \item $v$ is an $L$-stable model of $P$ iff $v$ is an $L$-stable model of $\Xi(P)$.
\end{itemize}

\end{restatable}

From Theorems \ref{t:pstable} and \ref{t:equivalence}, we see the $\nlp$ $P$ from Example \ref{ex:nlp} and the corresponding $\nadf$ $\Xi(P)$ from Example \ref{ex:xip} produce the same semantics. This result sheds light on the connections between $\adf$s and $\nlp$s. 
Until now, it was unclear if any $\adf$ semantics could capture a 3-valued one for $\nlp$s. 
Theorem \ref{t:equivalence} ensures the translation from $\nlp$ to $\adf$ in Definition~\ref{d:xip} is robust enough to guarantee at least the equivalence between any semantics based on partial stable models (at the $\nlp$ side) with any semantics based on complete models (at the $\adf$ side).

\section{Related Works}\label{s:related}

The relation between $\nlp$ and formal argumentation goes back to works such as \cite{prakken1997argument,simari1992mathematical,dung95acceptability}. In the sequel, we will describe previous attempts to translate $\adf$s to $\nlp$s (Subsection \ref{ss:adf-lp}) and $\nlp$s to $\adf$s (Subsection \ref{ss:lp-adf}) and the main connections between $\nlp$s and other argument-based frameworks such as Abstract Argumentation Frameworks ($\aaf$s) \cite{dung95acceptability} and Assumption-Based Argumentation (ABA) \cite{dung2009assumption} in Subsection \ref{ss:others}. Afterwards, we compare an extension of $\aaf$, called $\setaf$ \cite{nielsen2006generalization}, with $\nadf$.

\subsection{From $\adf$ to Logic Programming}\label{ss:adf-lp}

 As pointed out by \cite{strass2013approximating}, there is a direct translation from $\adf$s to $\nlp$s:
 
 \begin{mdef}[\protect \cite{strass2013approximating}]\label{d:pxi}
Let $\Xi = (S, L, C^\tr)$ be an $\adf$. Define the corresponding $\nlp$ $P(\Xi) = $ $\left\{s \leftarrow a_1, \ldots a_m, \right.$ $\left. \naf b_1, \ldots, \naf b_n \mid s \in S, \set{a_1, \ldots a_m} \hspace{-.1em} \in C^\tr_s \textit{ and } \set{b_1, \ldots, b_n} = \pr(s) -\set{a_1, \ldots, a_m} \right\}$.
 \end{mdef}
 
 Note the body of a rule for $s$ is satisfied by an interpretation $I$ whenever for some $R \subseteq C^\tr_s$, the statements in $R$ are $\tr$ in $I$ and the remaining parents of $s$ are $\fa$ in $I$.

\begin{example}\label{ex:pxi} The $\nlp$ $P(\Xi)$ corresponding to the $\adf$ $\Xi$ of Example \ref{ex:adf} is given by 
\[
P(\Xi) = \left\{
\begin{array}{lllll}
a \leftarrow \naf b \quad &  d \leftarrow \naf c \quad & c \leftarrow e, \naf b \quad b \leftarrow \naf a \quad  &  e \leftarrow \naf d
\end{array}
\right\}.
\]
\end{example}

An $\adf$ $\Xi$ and the corresponding $\nlp$ $P(\Xi)$ are equivalent under various well-known semantics \cite{strass2013approximating}. Indeed, the complete models, grounded models, preferred models and stable models of $\Xi$ correspond respectively to the partial stable models, grounded models, regular models and stable models of $P(\Xi)$. This result allows us to say $\adf$s are as expressive as $\nlp$s. From an $\nlp$ $P$, we obtain an $\adf$ $\Xi(P)$ via Definition \ref{d:xip}, and then again an $\nlp$ $P(\Xi(P))$ via Definition \ref{d:pxi}. Although $P$ and $P(\Xi(P))$ are equivalent according to the aforementioned semantics, it is not guaranteed $P = P(\Xi(P))$:

Recall the $\nlp$ $P$ in Example \ref{ex:nlp} and the corresponding $\nadf$ $\Xi(P)$ in Example \ref{ex:xip}. From $\Xi(P)$ via Definition \ref{d:pxi}, we obtain the $\nlp$ $P(\Xi(P))$ (note $P \neq P(\Xi(P))$):
\[ 
\begin{array}{llllll}
	b \leftarrow \naf a	& \qquad    a \leftarrow \naf b  & \qquad c \leftarrow \quad &	p \leftarrow \naf p\quad  & p \leftarrow \naf a\quad  & d \leftarrow. \\
\end{array}
\]

Similarly, from an $\adf$ $\Xi$, we can obtain the $\nlp$ $P(\Xi)$ (Definition \ref{d:pxi}), and then again an $\adf$ $\Xi(P(\Xi))$ (Definition \ref{d:xip}). As above, they will be equivalent according to the aforementioned semantics, however, it does not guarantee $\Xi = \Xi(P(\Xi))$.

Recall the $\adf$ $\Xi$ in Example \ref{ex:adf} and the corresponding $\nlp$ $P(\Xi)$ in Example \ref{ex:pxi}. From $P(\Xi)$ via Definition \ref{d:xip}, we obtain the $\adf$ $\Xi(P(\Xi))$ (note $\Xi \neq \Xi(P(\Xi))$):
\[ a [\neg b] \qquad b[\neg a] \qquad c [\neg b \wedge \neg d] \qquad d [\neg c] \qquad e [\neg d]. \]

\subsection{From Logic Programming to $\adf$}\label{ss:lp-adf}

As we have mentioned, previous attempts to identify a semantics for $\adf$s equivalent to a 3-valued semantics for $\nlp$s have failed \cite{brewka2010abstract,strass2013approximating}.

\begin{mdef}[\protect \cite{brewka2010abstract}]\label{d:xip2}
Let $P$ be an $\nlp$ over a set $A$ of atoms. 
Define an $\adf$, $\Xi_2(P) = (A, L, C^\tr)$, in which 
\begin{itemize}
    \item $L\hspace{-.2em} = \hspace{-.2em} \set{(c,a)\hspace{-.2em} \mid\hspace{-.2em} a \leftarrow a_1, \ldots, a_m, \naf b_1, \ldots, \naf b_n\hspace{-.2em} \in P\hspace{-.2em} \textit{ and } c \in\hspace{-.08em} \set{a_1, \ldots, a_m, b_1, \ldots, b_n}}$;
    \item For each $a \in A$, $C^\tr_a = \{B \in \pr(a) \mid a \leftarrow a_1, \ldots, a_m, \naf b_1, \ldots, \naf b_n \in P, \{a_1, \ldots, a_m \} \subseteq B, \set{b_1, \ldots, b_n} \cap B = \emptyset \}$.
\end{itemize}
Alternatively, we could define the acceptance condition of each $a \in A$ as 
\begin{align}\label{eq:xip2}
\varphi_a \equiv \bigvee_{a \leftarrow a_1, \ldots, a_m, \naf b_1, \ldots, \naf b_n \in P} \left( a_1 \wedge \cdots \wedge a_m \wedge \neg b_1 \wedge \cdots \wedge \neg b_n \right).
\end{align}
\end{mdef}

 As noted in \cite{strass2013approximating}, by Definition \ref{d:xip2}, the $\nlp$s,
$P_1 = \{c \leftarrow; b \leftarrow \naf b;$ $a \leftarrow b; a \leftarrow c\}$ and $P_2 = \set{c \leftarrow; b \leftarrow \naf b; a \leftarrow b, \naf c; a \leftarrow c, \naf b; a \leftarrow b, c}$
produce the same $\adf$s: ($\Xi_2(P_1) = \Xi_2(P_2)$). For any $s \in A$, its corresponding acceptance condition is $c [\tr] \qquad b[\neg b] \qquad a [b \vee c]\footnote{By Equation \ref{eq:xip2}, for $\Xi_2(P)$, we have $\varphi_a \equiv (b \wedge \neg c) \vee (\neg b \wedge c) \vee (b \wedge c) \equiv b \vee c$.}$. But the unique partial stable model (\textit{PSM}) of $P_1$ is  $\set{a, c}$, whereas $\set{c}$ is the unique \textit{PSM} of $P_2$. 
Hence, this translation is inadequate to distinguish these two non-equivalent programs, according to \textit{PSM}s. 
In contradistinction, our translation works accordingly and produces respectively the $\adf$s below, which has the same semantics as their corresponding original programs: $\Xi(P_1)$ is given by $ c [\tr] \qquad b[\neg b] \qquad a [\tr \vee \neg b] $, and $\Xi(P_2)$ is given by 
$c [\tr] \qquad b[\neg b] \qquad a [(\neg b \vee \neg c) \vee \neg b]$.
However, when restricting to the class of $\nlp$s where each rule is as $a \leftarrow \naf b_1, \ldots, \naf b_m$, $m \geq 0$, the translation of Definition \ref{d:xip2} coincides with the translation of Definition \ref{d:xip} and is robust enough to capture 3-valued semantics as \textit{PSM} and well-founded models. 

\begin{restatable}{prop}{xipII}\label{p:xip2}
Let $P$ be an $\nlp$, where each rule is either a fact or its body has only default literals as in $a \leftarrow \naf b_1, \ldots, \naf b_n$. 
Let $\Xi(P)$ be the $\adf$ obtained from $P$ via Definition~\ref{d:xip} and $\Xi_2(P)$ the $\adf$ obtained from $P$ via Definition~\ref{d:xip2}. Then $\Xi(P) = \Xi_2(P)$.
\end{restatable}

This result shows how Definition \ref{d:xip2} could be employed to capture 3-valued semantics as \textit{PSM}s: firstly, one could take an $\nlp$ $P$ and apply any program transformation (preserving \textit{PSM}s) that transforms an $\nlp$ into one as that of Proposition \ref{p:xip2}\footnote{See \cite{brass1995characterizations} for some program transformations.}. Then, one could apply the translation in Definition \ref{d:xip2} to the resulting program (say $P'$) to obtain $\Xi_2(P')$. From Proposition \ref{p:xip2}, it holds $P$ and $\Xi_2(P')$ have the same \textit{PSM}s.

\subsection{On the connections between Logic Programming and Argumentation}\label{ss:others}

Logic programming has long served as an inspiration for argumentation theory. Indeed, one can see the seminal work of Dung \cite{dung95acceptability} on Abstract Argumentation Frameworks ($\aaf$s) as an abstraction of some aspects of logic programming. 
In \cite{caminada15equivalence}, the authors pointed out that the translation from logic programming to these frameworks described in \cite{wu2009complete} is able to account for the equivalences between Partial Stable Models, Well-Founded Models, Regular Models, Stable Models Semantics for $\nlp$s and respectively Complete Models, grounded models, Preferred Models, Stable Models for $\aaf$s. 
However, unlike we have done for $\adf$s, they have showed that, with their proposed translation from $\nlp$s to $\aaf$s, there cannot be a semantics for $\aaf$s equivalent to $L$-Stable Semantics for $\nlp$s.

When translating $\aaf$s to $\nlp$s, the connection between their semantics is stronger than when translating in the opposite direction as for any of the mentioned semantics for $\aaf$s; there exists an equivalent semantics for $\nlp$s \cite{caminada15equivalence}.

In \cite{caminada2017equivalence}, the authors showed how to translate Assumption-Based Argumentation (\textit{ABA}) \cite{bondarenko1997abstract,dung2009assumption,toni2014tutorial} to $\nlp$s and how this translation can be reapplied for a reverse translation from $\nlp$s to \textit{ABA}. 
Curiously, the problematic direction here is from \textit{ABA} to $\nlp$. 
In \cite{caminada2017equivalence}, they have showed that with their proposed translation, there cannot be a semantics for $\nlp$s equivalent to the the semi-stable semantics \cite{caminada2015difference,schulz2015logic} for \textit{ABA}.

\subsection{A Comparison between $\setaf$ and $\nadf$}\label{ss:setaf}

In \cite{nielsen2006generalization}, they proposed an extension of Dung's Abstract Argumentation Frameworks ($\aaf$s) to allow joint attacks on arguments. The resulting framework, called $\setaf$, is displayed below:

\begin{mdef}[\protect \cite{nielsen2006generalization}]\label{d:setaf}
A framework with sets of attacking arguments ($\setaf$) is a pair $\mathit{SF} = (A, R)$, where $A$ is the set of arguments and $R \subseteq (2^A - \emptyset) \times A$
is the attack relation.
\end{mdef}

In an $\aaf$, the unique relation between arguments is given by the attack relation, where an (individual) argument attacks another. In a $\setaf$ (as well as in an $\nadf$), the novelty is that a set of arguments can attack an argument. For a translation from $\setaf$ to $\adf$ refer to \cite{polberg2016understanding}:

\textbf{Translation.} Let $\mathit{SF} = (A, R)$ be a $\setaf$. 
The $\adf$ corresponding to $\mathit SF$ is $\mathit{DF}^{\mathit{SF}} = (A, L, C)$, where  $L = \{(x, y) \mid \exists X \subseteq A \textit{ such that } x \in X \textit{ and } (X, y) \in R \}$, $C = \set{C_a}, {a \in A}$ and every $C_a$ gets constructed in the following way: for every $B \subseteq \pr(a)$, if $\exists (X_i, a) \in R$ such that $X_i \subseteq B$, then $C_a(B) = \fa$; otherwise, $C_a(B) = \tr$.

The following result is immediate:

\begin{restatable}{prop}{setaff}\label{p:setaf}
Let 
$\mathit{SF} = (A, R)$ 
be a $\setaf$ and 
$\mathit{DF}^{\mathit{SF}} = (A, L, C)$ 
be the corresponding $\adf$.  
Then, $\mathit{DF}^{\mathit{SF}}$ is an $\nadf$.
\end{restatable}

On the other hand, not every $\nadf$ will correspond to a $\setaf$ according to the translation above. A noticeable difference between them is that for every argument $a \in A$ in a $\setaf$ $\mathit{SF} = (A, R)$, it holds $(\emptyset, a) \not\in R$. 
Then, for every statement $s$ in the corresponding $\mathit{DF}^{\mathit{SF}}$, it holds $C_s(\emptyset) = \tr$, while $C_s(\emptyset) = \fa$ is allowed in $\nadf$. 
Indeed, when $C_s(\emptyset) = \fa$ in an $\nadf$, we have $C_s(R) = \fa$ for every $R \subseteq \pr(s)$, i.e., $\varphi_s \equiv \fa$.

\section{Conclusions and Future Works}\label{s:conclusions}

In this paper, we have investigated the connections between Abstract Dialectical Frameworks ($\adf$s) and Normal Logic Programs ($\nlp$s). 
Unlike previous works \cite{brewka2010abstract,strass2013approximating}, we have provided a translation from $\nlp$s to $\adf$s robust enough to capture the equivalence between several frameworks for these formalisms, including 3-valued semantics. 
In particular, after resorting to our translation, we have proved the equivalence between partial stable models, well-founded models, regular models, stable models semantics for $\nlp$s and respectively complete models, grounded models, preferred models, stable models for $\adf$s. 

Curiously, we have obtained these equivalence results by translating an $\nlp$ into a fragment of $\adf$, called Attacking Dialectical Frameworks ($\nadf$), in which the unique relation involving statements is the attack relation. A distinguishing aspect of our translation when compared with related works as \cite{caminada15equivalence, strass2013approximating} is that it is made in two steps: in the first step each $\nlp$ rule is translated into a substatement, and then, substatements corresponding to rules with the same head are gathered to constitute a unique statement. With this procedure, our intention is to simulate the semantics for $\nlp$s, where the truth-value of an atom $b$ is the disjunction of the truth-values of the bodies of the rules whose head is $b$. Besides, we have defined a new semantics for $\nadf$, called $L$-Stable, and showed it is equivalent to the $L$-Stable Semantics (defined in \cite{eiter97partial}) for $\nlp$s. 

An essential element to define these semantics for $\adf$ is $\Gamma_D$, a kind of immediate consequences operator. 
When restricted to $\nadf$, we have proved $\Gamma_D$ is equivalent to a noticeably simpler version. 
Indeed, owing to this simplicity, verifying whether a given labelling is complete is of complexity $P$, whereas this verification problem is DP-complete for $\adf$ \cite{brewka2013abstract}.
This is a promising result as it might also mean the complexity of many reasoning tasks on $\nadf$s may have the same complexity as standard Dung's Abstract Argumentation Frameworks \cite{dung95acceptability}.

As future work, we intend to complete a thorough investigation of the connections between $\adf$s and $\nadf$s. 
Regarding the equivalences between $\nlp$ and $\nadf$, one can claim that $\nadf$s are as general as $\adf$s, and the attack relation suffices to express these relations involving statements in $\adf$s. 
Given the results unveiled in the current paper, we also envisage  unfolding the connections between $\nlp$s and $\setaf$s \cite{nielsen2006generalization}, an extension of Dung's Abstract Argumentation Frameworks to allow joint attacks on arguments. We expect that there are various correspondences between their semantics.

\bibliographystyle{acmtrans}
\bibliography{iclp}

\begin{thebibliography}{}

\bibitem[\protect\citeauthoryear{Bondarenko, Dung, Kowalski, and
  Toni}{Bondarenko et~al\mbox{.}}{1997}]{bondarenko1997abstract}
{\sc Bondarenko, A.}, {\sc Dung, P.~M.}, {\sc Kowalski, R.~A.}, {\sc and} {\sc
  Toni, F.} 1997.
\newblock An abstract, argumentation-theoretic approach to default reasoning.
\newblock {\em Art. Intelligence\/}~{\em 93,\/}~1-2, 63--101.

\bibitem[\protect\citeauthoryear{Brass and Dix}{Brass and
  Dix}{1995}]{brass1995characterizations}
{\sc Brass, S.} {\sc and} {\sc Dix, J.} 1995.
\newblock Characterizations of the stable semantics by partial evaluation.
\newblock In {\em International Conf. on Logic Programming and Nonmonotonic
  Reasoning}. Springer, 85--98.

\bibitem[\protect\citeauthoryear{Brewka, Ellmauthaler, Strass, Wallner, and
  Woltran}{Brewka et~al\mbox{.}}{2013}]{brewka2013abstract}
{\sc Brewka, G.}, {\sc Ellmauthaler, S.}, {\sc Strass, H.}, {\sc Wallner,
  J.~P.}, {\sc and} {\sc Woltran, S.} 2013.
\newblock Abstract dialectical frameworks revisited.
\newblock In {\em Proceedings of the Twenty-Third international joint
  conference on Artificial Intelligence}. AAAI Press, 803--809.

\bibitem[\protect\citeauthoryear{Brewka and Woltran}{Brewka and
  Woltran}{2010}]{brewka2010abstract}
{\sc Brewka, G.} {\sc and} {\sc Woltran, S.} 2010.
\newblock Abstract dialectical frameworks.
\newblock In {\em Twelfth International Conf. on the Principles of Knowledge
  Representation and Reasoning}. AAAI Press, 102--111.

\bibitem[\protect\citeauthoryear{Buss}{Buss}{1987}]{buss1987boolean}
{\sc Buss, S.~R.} 1987.
\newblock The boolean formula value problem is in alogtime.
\newblock In {\em Proceedings of the nineteenth annual ACM symposium on Theory
  of computing}. ACM, 123--131.

\bibitem[\protect\citeauthoryear{Caminada}{Caminada}{2006}]{caminada2006semi}
{\sc Caminada, M.} 2006.
\newblock Semi-stable semantics.
\newblock {\em 1st International Conference on Computational Models of Argument
  (COMMA)\/}~{\em 144}, 121--130.

\bibitem[\protect\citeauthoryear{Caminada, S{\'a}, Alc{\^a}ntara, and
  Dvo{\v{r}}{\'a}k}{Caminada et~al\mbox{.}}{2015a}]{caminada15equivalence}
{\sc Caminada, M.}, {\sc S{\'a}, S.}, {\sc Alc{\^a}ntara, J.}, {\sc and} {\sc
  Dvo{\v{r}}{\'a}k, W.} 2015a.
\newblock On the equivalence between logic programming semantics and
  argumentation semantics.
\newblock {\em International Journal of Approximate Reasoning\/}~{\em 58},
  87--111.

\bibitem[\protect\citeauthoryear{Caminada and Schulz}{Caminada and
  Schulz}{2017}]{caminada2017equivalence}
{\sc Caminada, M.} {\sc and} {\sc Schulz, C.} 2017.
\newblock On the equivalence between assumption-based argumentation and logic
  programming.
\newblock {\em Journal of Artificial Intelligence Research\/}~{\em 60},
  779--825.

\bibitem[\protect\citeauthoryear{Caminada, S{\'a}, Alc{\^a}ntara, and
  Dvo{\v{r}}{\'a}k}{Caminada et~al\mbox{.}}{2015b}]{caminada2015difference}
{\sc Caminada, M. W.~A.}, {\sc S{\'a}, S.}, {\sc Alc{\^a}ntara, J.}, {\sc and}
  {\sc Dvo{\v{r}}{\'a}k, W.} 2015b.
\newblock On the difference between assumption-based argumentation and abstract
  argumentation.
\newblock {\em IfCoLog Journal of Logics and their Applications\/}.

\bibitem[\protect\citeauthoryear{Dung}{Dung}{1995}]{dung95acceptability}
{\sc Dung, P.} 1995.
\newblock On the acceptability of arguments and its fundamental role in
  nonmonotonic reasoning, logic programming and $n$-person games.
\newblock {\em Artificial Intelligence\/}~{\em 77}, 321--357.

\bibitem[\protect\citeauthoryear{Dung, Kowalski, and Toni}{Dung
  et~al\mbox{.}}{2009}]{dung2009assumption}
{\sc Dung, P.~M.}, {\sc Kowalski, R.~A.}, {\sc and} {\sc Toni, F.} 2009.
\newblock Assumption-based argumentation.
\newblock In {\em Argumentation in artificial intelligence}. Springer,
  199--218.

\bibitem[\protect\citeauthoryear{Eiter, Leone, and Sacc\'{a}}{Eiter
  et~al\mbox{.}}{1997}]{eiter97partial}
{\sc Eiter, T.}, {\sc Leone, N.}, {\sc and} {\sc Sacc\'{a}, D.} 1997.
\newblock On the partial semantics for disjunctive deductive databases.
\newblock {\em Ann. Math. Artif. Intell.\/}~{\em 19,\/}~1-2, 59--96.

\bibitem[\protect\citeauthoryear{Ellmauthaler}{Ellmauthaler}{2012}]{ellmauthaler2012}
{\sc Ellmauthaler, S.} 2012.
\newblock Abstract {D}ialectical {F}rameworks: {P}roperties, {C}omplexity, and
  {I}mplementation.
\newblock M.S.\ thesis, Technische Universit\"at Wien, Institut f\"ur
  Informationssysteme.

\bibitem[\protect\citeauthoryear{Gelfond and Lifschitz}{Gelfond and
  Lifschitz}{1988}]{gelfond1988stable}
{\sc Gelfond, M.} {\sc and} {\sc Lifschitz, V.} 1988.
\newblock The stable model semantics for logic programming.
\newblock In {\em Proc. of the 5th International Conference on Logic
  Programming (ICLP)}. Vol.~88. 1070--1080.

\bibitem[\protect\citeauthoryear{Kleene, de~Bruijn, de~Groot, and
  Zaanen}{Kleene et~al\mbox{.}}{1952}]{kleene1952introduction}
{\sc Kleene, S.~C.}, {\sc de~Bruijn, N.}, {\sc de~Groot, J.}, {\sc and} {\sc
  Zaanen, A.~C.} 1952.
\newblock {\em Introduction to metamathematics}. Vol. 483.
\newblock van Nostrand New York.

\bibitem[\protect\citeauthoryear{Nielsen and Parsons}{Nielsen and
  Parsons}{2006}]{nielsen2006generalization}
{\sc Nielsen, S.~H.} {\sc and} {\sc Parsons, S.} 2006.
\newblock A generalization of {D}ung’s abstract framework for argumentation:
  Arguing with sets of attacking arguments.
\newblock In {\em International Workshop on Argumentation in Multi-Agent
  Systems}. Springer, 54--73.

\bibitem[\protect\citeauthoryear{Polberg}{Polberg}{2016}]{polberg2016understanding}
{\sc Polberg, S.} 2016.
\newblock Understanding the abstract dialectical framework.
\newblock In {\em European Conference on Logics in Artificial Intelligence}.
  Springer, 430--446.

\bibitem[\protect\citeauthoryear{Prakken and Sartor}{Prakken and
  Sartor}{1997}]{prakken1997argument}
{\sc Prakken, H.} {\sc and} {\sc Sartor, G.} 1997.
\newblock Argument-based extended logic programming with defeasible priorities.
\newblock {\em Journal of applied non-classical logics\/}~{\em 7,\/}~1-2,
  25--75.

\bibitem[\protect\citeauthoryear{Przymusinski}{Przymusinski}{1990}]{przymusinski90well-founded}
{\sc Przymusinski, T.~C.} 1990.
\newblock The well-founded semantics coincides with the three-valued stable
  semantics.
\newblock {\em Fundamenta Informaticae\/}~{\em 13,\/}~4, 445--463.

\bibitem[\protect\citeauthoryear{Schulz and Toni}{Schulz and
  Toni}{2015}]{schulz2015logic}
{\sc Schulz, C.} {\sc and} {\sc Toni, F.} 2015.
\newblock Logic programming in assumption-based argumentation
  revisited-semantics and graphical representation.
\newblock In {\em 29th AAAI Conf. on Art. Intelligence}.

\bibitem[\protect\citeauthoryear{Simari and Loui}{Simari and
  Loui}{1992}]{simari1992mathematical}
{\sc Simari, G.~R.} {\sc and} {\sc Loui, R.~P.} 1992.
\newblock A mathematical treatment of defeasible reasoning and its
  implementation.
\newblock {\em Artificial intelligence\/}~{\em 53,\/}~2-3, 125--157.

\bibitem[\protect\citeauthoryear{Strass}{Strass}{2013}]{strass2013approximating}
{\sc Strass, H.} 2013.
\newblock Approximating operators and semantics for abstract dialectical
  frameworks.
\newblock {\em Artificial Intelligence\/}~{\em 205}, 39--70.

\bibitem[\protect\citeauthoryear{Toni}{Toni}{2014}]{toni2014tutorial}
{\sc Toni, F.} 2014.
\newblock A tutorial on assumption-based argumentation.
\newblock {\em Argument \& Computation\/}~{\em 5,\/}~1, 89--117.

\bibitem[\protect\citeauthoryear{Wu, Caminada, and Gabbay}{Wu
  et~al\mbox{.}}{2009}]{wu2009complete}
{\sc Wu, Y.}, {\sc Caminada, M.}, {\sc and} {\sc Gabbay, D.~M.} 2009.
\newblock Complete extensions in argumentation coincide with 3-valued stable
  models in logic programming.
\newblock {\em Studia logica\/}~{\em 93,\/}~2-3, 383.

\end{thebibliography}


\begin{thebibliography}{}

\bibitem[\protect\citeauthoryear{Butcher}{Butcher}{1981}]{Butcher}
{\sc Butcher, J.} 1981.
\newblock {\em Copy-editing: The Cambridge Handbook}.
\newblock Cambridge University Press.

\bibitem[\protect\citeauthoryear{{{C}adence {R}esearch {S}ystems}}{{{C}adence
  {R}esearch {S}ystems}}{1994}]{crs:chez}
{\sc {{C}adence {R}esearch {S}ystems}}. 1994.
\newblock {\it {C}hez} {S}cheme {R}eference {M}anual.

\bibitem[\protect\citeauthoryear{Cameron and Ito}{Cameron and
  Ito}{1984}]{ci:gramps}
{\sc Cameron, R.~D.} {\sc and} {\sc Ito, M.~R.} 1984.
\newblock Grammar-based definition of metaprogramming systems.
\newblock {\em {ACM} Transactions on Programming Languages and Systems\/}~{\em
  6,\/}~1 (Jan.), 20--54.

\bibitem[\protect\citeauthoryear{Grossman}{Grossman}{1982}]{Chicago}
{\sc Grossman, J.}, Ed. 1982.
\newblock {\em The Chicago Manual of Style}.
\newblock University of Chicago Press.

\bibitem[\protect\citeauthoryear{Lamport}{Lamport}{1986}]{LaTeX}
{\sc Lamport, L.} 1986.
\newblock {\em \LaTeX: A Document Preparation System\/}, 2 ed.
\newblock Addison-Wesley, New York.

\end{thebibliography}

\appendix

\section{Proofs of Theorems} \label{app-proofs}

\subsection{Theorems and Proofs from Section \ref{s:attacking}:}

\nnadf*

\begin{proof}

According to Equation (\ref{eq:acceptance}), $\varphi_s \equiv \varphi_1 = \bigvee_{R \in C^\tr_s} \left( \bigwedge_{a \in R} a \wedge \bigwedge_{b \in \pr(s) - R} \neg b \right)$. Let $ \varphi_2 = \bigvee_{R \in C^\mathit{max}_s}  \bigwedge_{b \in \pr(s) - R} \neg b $. We will show $\varphi_1 \equiv \varphi_2$, i.e., for any 2-valued interpretation $v$, $v(\varphi_1) = v(\varphi_2)$:

\begin{itemize}
    \item If $v(\varphi_1) = \tr$, then there exists $R \in C^\tr_s$ such that for all $a \in R$, $v(a) = \tr$ and for all $b \in \pr(s) - R$, $v(b) = \fa$. As there exists $R' \in C^\mathit{max}_s$ such that $R \subseteq R'$, we obtain for all $b \in \pr(s) - R'$, $v(b) = \fa$. Thus, $v(\varphi_2) = \tr$.
    \item If $v(\varphi_1) = \fa$, then for each $R \in C^\tr_s$ there exists $a \in R$ such that $v(a) = \fa$ or there exists $b \in \pr(s) - R$ such that $v(b) = \tr$. In particular, for each $R \in C^\mathit{max}_s$ there exists $a \in R$ such that $v(a) = \fa$ or there exists $b \in \pr(s) - R$ such that $v(b) = \tr$, and\footnote{As $D$ is an $\nadf$, for each $R \in C^\mathit{max}_s$, for each $R' \subseteq R$, we have $R' \in C^\tr_s$.} there exists $b \in \pr(s) - R'$ such that $v(b) = \tr$, in which $R' = R - \set{a \in R \mid v(a) = \fa}$. But then for each $R \in C^\mathit{max}_s$ there exists $b \in \pr(s) - R$ such that $v(b) = \tr$. Thus, $v(\varphi_2) = \fa$.

\comments{    \item If $v(\varphi_1) = \un$, then for each $R \in C^\tr_s$, there exists $a \in R$ such that $v(a) \neq \tr$ or there exists $b \in \pr(s) - R$ such that $v(b) \neq \fa$, and there exists $R' \in C^\tr_s$ such that for each $a \in R'$, $v(a) \neq \fa$ and for each $b \in \pr(s) - R'$, $v(b) \neq \tr$. This means 
    
    \begin{itemize}
        \item for each $R \in C^\mathit{max}_s$, there exists $a \in R$ such that $v(a) \neq \tr$ or there exists $b \in \pr(s) - R$ such that $v(b) \neq \fa$, and (as $D$ is an $\nadf$) there exists $b \in \pr(s) - R_1$ such that $v(b) \neq \fa$, in which $R_1 = R - \set{a \in R \mid v(a) \neq \tr}$, and 
        \item there exists $R' \in C^\mathit{max}_s$ such that for each $b \in \pr(s) - R'$, $v(b) \neq \tr$. 
    \end{itemize}
    
    But then for each $R \in C^\mathit{max}_s$, there exists $b \in \pr(s) - R$ such that $v(b) \neq \fa$, and there exists $R' \in C^\mathit{max}_s$ such that for each $b \in \pr(s) - R'$, $v(b) \neq \tr$. Thus, $v(\varphi_2) = \un$.
} 
\end{itemize}

\end{proof}

\redundant*

\begin{proof}

$(\Rightarrow)$

If $(r,s) \in L$ is a redundant link, then, in particular, it is a supporting link, i.e., for every $R \subseteq \pr(s)$, we have if $R \in C^\tr_s$, then $(R \cup \set{r}) \in C^\tr_s$.

By absurd, suppose there exists $R \in C^\mathit{max}_s$ such that $r \not\in R$. This means $R \in C^\tr_s$. But then we obtain $(R \cup \set{r}) \in C^\tr_s$. It is an absurd as $R \in C^\mathit{max}_s$.

\noindent$(\Leftarrow)$

Assume for any $R \in C^\mathit{max}_s$, we have $r \in R$. By absurd, suppose $(r,s) \in L$ is not redundant. Then there exists $R' \subseteq \pr(s)$ such that $C_s(R') = \tr$ and $C_s(R' \cup \set{r}) = \fa$.

As $r \in R$ for any $R \in C^\mathit{max}_s$, there exists $R'' \in C^\mathit{max}_s$ such that $R' \cup \set{r} \subseteq R''$ and $C_s(R'') = \tr$. But then, as any link in $L$ is attacking, we obtain $C_s(R' \cup \set{r}) = \tr$. An absurd.
\end{proof}

\redundantII*

\begin{proof}
The result is straightforward: from Theorem \ref{t:redundant}, we know $(r,s) \in L$ is a redundant link iff for any $R \in C^\mathit{max}_s$, we have $r \in R$ iff $\neg r$ does not appear in $\displaystyle \bigvee_{R \in C^\mathit{max}_s}  \bigwedge_{b \in \pr(s) - R} \neg b$ iff $(r,s) \not\in L'$.
\end{proof}

\redundantIII*

\begin{proof}
The proof follows from the definition of $\nadf$, a property of Power Sets and the Principle of Inclusion and Exclusion (PIE).

In $D$, for every $s \in S$ and $M \subseteq par(s)$, if $C_s(M) = \tr$, then $C_s(M') = \tr$ for every $M' \subseteq M$ (Definition \ref{def:nadf}). Then $C^\tr_s = \set{ S \subseteq R \mid R \in C^\mathit{max}_s }$ = $\bigcup \set{ \powerset(R) \mid R \in C^\mathit{max}_s }$, where $C^\mathit{max}_s = \left\{R \in C^\tr_s \mid \textit{there is} \right.$ $\left. \textit{no } R' \in C^\tr_s \textit{ such that } R \subset R' \right\}$ and $\powerset(R)$ denotes the power set of $R$.


Given a set $S$, we have $|\powerset(S)| = 2^{|S|}$ and that, for each $r \in S$, $r$ is an element of $\frac{2^{|S|}}{2}$ subsets of $S$, i.e., of precisely half the subsets of $S$. Then if $r \in S \cap T$, we have that $r$ is an element of $\frac{2^{|S|}}{2}$ subsets of $S$, $\frac{2^{|T|}}{2}$ subsets of $T$ and $\frac{2^{|S \cap T|}}{2}$ subsets of $S \cap T$. PIE ensures that $|\powerset(S) \cup \powerset(T)| = |\powerset(S)| + |\powerset(T)| - |\powerset(S) \cap \powerset(T)|$, which, because $\powerset(S \cap T) = \powerset(S) \cap \powerset(T)$, leads to $|\powerset(S) \cup \powerset(T)| = |\powerset(S)| + |\powerset(T)| - |\powerset(S \cap T)|$. That is, if $r \in S \cap T$, then $|\powerset(S) \cup \powerset(T)| = 2^{|S|} + 2^{|T|} - 2^{|S \cap T|}$ and $r$ is an element of $\frac{2^{|S|}}{2} + \frac{2^{|T|}}{2} - \frac{2^{|S \cap T|}}{2} = \frac{ |\powerset(S) \cup \powerset(T)|}{2}$ sets in $\powerset(S) \cup \powerset(T)$. By extension of PIE, if $r \in \bigcap\set{S_1,\ldots,S_n}$, then $r$ is an element of $\frac{ |\bigcup\set{\powerset(S_1),\ldots,\powerset(S_n)}| }{2}$ sets in $\bigcup\set{\powerset(S_1),\ldots,\powerset(S_n)}$.

Let $(r,s)$ be a redundant link, then, for all $R \in C^\mathit{max}_s$, we have $r \in R$ (Theorem \ref{t:redundant}), i.e., $r \in \bigcap{C^\mathit{max}_s}$. Then $r$ is an element of $\frac{|\bigcup \set{ \powerset(R)\ \mid\ R \in C^\mathit{max}_s\ and\ r \in R }|}{2} = \frac{|C^\tr_s|}{2}$ sets in $\bigcup \set{ \powerset(R)\ \mid\ R \in C^\mathit{max}_s\ and\ r \in R } = C^\tr_s$, i.e., $|C^\tr_s(r)| = \frac{|C^\tr_s|}{2}$.
\end{proof}

\linear*

\begin{proof}
Because $|C^\tr_s(r)| = \frac{|C^\tr_s|}{2}$, where $C^\tr_s(r) = \set{ R \in C^\tr_s \mid r \in R }$, to find if $(r,s)$ is a redundant link, it suffices to check for each $R \in C^\tr_s$, if $r \in R$. For each $R \in C^\tr_s$, checking if $r \in R$ can be done by checking, for each $s \in R$, if $s = r$. Clearly, each $R \in C^\tr_s$ has at most $k = max \set{ |R| \mid R \in \bigcup C^\mathit{max}_s  }$ elements. Because $C^\mathit{max}_s \subset C^\tr_s$ and $C^\tr_s$ is subset-complete, we have $|C^\tr_s| \geq 2^k$. Then $k$ is $O(ln|C^\tr_s|)$, which means that deciding if a link $(r,s) \in L$ is redundant is $O(|C^\tr_s|.ln(|C^\tr_s|))$.
\end{proof}

\gammaomega*

\begin{proof}
For each $s \in S$, let $\varphi_s$ be
\[ \bigvee_{R \in C^\mathit{max}_s}  \bigwedge_{b \in \pr(s) - R} \neg b \]
It is enough to prove for each $s \in S$,  $v(\varphi_s) = \bigsqcap \set{w(\varphi_s) \mid w \in [v]_2}$,  where $[v]_2 = \set{w \mid w \textit{ is two-valued and } v \leq_i w}$. We have three possibilities:

\begin{itemize}
    \item $v(\varphi_s) = \tr$ iff there exists $R \in C^\mathit{max}_s$ such that for each $b \in \pr(s) - R$, $v(b) = \fa$ iff there exists $R \in C^\mathit{max}_s$ such that for each $b \in \pr(s) - R$,  for each $w \in [v]_2$,  $w(b) = \fa$ iff for each $w \in [v]_2$, $w(\varphi_s) = \tr$ iff $\bigsqcap \set{w(\varphi_s) \mid w \in [v]_2} = \tr$.
    \item $v(\varphi_s) = \fa$ iff for each $R \in C^\mathit{max}_s$, there exists $b \in \pr(s) - R$ such that $v(b) = \tr$ iff for each $w \in [v]_2$, for each $R \in C^\mathit{max}_s$, there exists $b \in \pr(s) - R$ such that $w(b) = \tr$ iff for every $w \in [v]_2$, $w(\varphi_s) = \fa$ iff $\bigsqcap \set{w(\varphi_s) \mid w \in [v]_2} = \fa$.
    \item $v(\varphi_s) = \un$, then for each $R \in C^\mathit{max}_s$, there exists $b \in \pr(s) - R$ such that $v(b) \in \set{\tr, \un}$ and there exists $R \in C^\mathit{max}_s$ such that for each $b \in \pr(s) - R$, it holds $v(b) \in \set{\fa, \un}$. Hence, 
    \begin{itemize}
        \item there exists $w \in [v]_2$ such that for each $R \in C^\mathit{max}_s$, there exists $b \in \pr(s) - R$ such that $w(b) = \tr$. This means there exists $w \in [v]_2$ such that $w(\varphi_s) = \fa$;
        \item there exists $w' \in [v]_2$, there exists $R \in C^\mathit{max}_s$ such that for each $b \in \pr(s) - R$, it holds $w'(b) = \fa$. This means there exists $w \in [v]_2$ such that $w(\varphi_s) = \tr$.
    \end{itemize}

But then we have $\bigsqcap \set{w(\varphi_s) \mid w \in [v]_2} = \un$.
\end{itemize}
\end{proof}

\stable*

\begin{proof}

$(\Rightarrow)$
Let $v$ be a stable model of $D$. It is trivial $v$ is a complete model of $D$ as every stable model is a complete model.

\noindent$(\Leftarrow)$

Let $v$ be a 2-valued complete model of $D$. We will show $v$ is a stable model of $D$, i.e., $v$ is a grounded model of $D^v = (E_v, L^v, C^v)$, in which $E_v = \set{s \in S \mid v(s) = \tr}$, $L^v = L \cap (E_v \times E_v)$ and for every $s \in E_v$, we set $\varphi_s^v = \varphi_s[b \slash \fa : v(b) = \fa]$.

As $v$ is a complete model of $D$, if $v(s) = \tr$, then $v(\varphi_s) = v(\bigvee_{R \in C^\mathit{max}_s}  \bigwedge_{b \in \pr(s) - R} \neg b) = \tr$. This means there exists $R \in C^\mathit{max}_s$ such that for each $b \in \pr(s) - R$, $v(b) = \fa$. Thus, for each $s \in E_v$, $\varphi_s^v \equiv \tr$. As consequence, $E_v$ is the grounded extension of $D^v$, i.e., $v$ is a stable model of $D$.
\end{proof}

\comments{
\completenadf*

\begin{proof}
From Definition \ref{d:admissible} and Theorem \ref{t:gammaomega}, we know a three-valued interpretation $v$ is a complete model of $D$ iff $\Omega_D(v) = v$. This means for each $s \in S$ we have $v(s) = v(\varphi_s)$. For each $s \in S$, we have three possibilities:

\begin{itemize}
    \item If $v(s) = \tr$, then $v(\varphi_s) = \tr$. This means  there exists $K \in C^\mathit{max}_s$ such that for each $b \in \pr(s) - K$, $v(b) = \fa$, i.e., there exists $K \in C^\tr_s$ such that for each $b \in \pr(s) - K$, $v(b) = \fa$;
    \item If $v(s) = \fa$, then $v(\varphi_s) = \fa$. This means for each $K \in C^\mathit{max}_s$, there exists $b \in \pr(s) - K$ such that $v(b) = \tr$, i.e., for each $K \in C^\tr_s$, there exists $b \in \pr(s) - K$ such that $v(b) = \tr$;
    \item If $v(s) = \un$, then $v(\varphi_s) = \un$. This means then for each $K \in C^\mathit{max}_s$, there exists $b \in \pr(s) - K$ such that $v(b) \in \set{\tr, \un}$ and there exists $K \in C^\mathit{max}_s$ such that for each $b \in \pr(s) - K$, it holds $v(b) \in \set{\fa, \un}$, i.e., for each $K \in C^\tr_s$, there exists $b \in \pr(s) - K$ such that $v(b) \neq \fa$ and there exists $K \in C^\tr_s$ such that for each $b \in \pr(s) - K$, it holds $v(b) \neq \tr$.
\end{itemize}
\end{proof}
}

\subsection{Theorems and Proofs from Section \ref{s:equiv}:}

\xipnadf*

\begin{proof}
Let $\Xi(P) = (A, L, C^\tr)$ be the $\adf$ corresponding to the $\nlp$ $P$ over a set of atoms $A$. By absurd, suppose $\Xi(P)$ is not an $\nadf$. This means there exists a link $(b,a) \in L$ for which some $R \subseteq \pr(a)$ we have $C_a(R) = \fa$ and $C_a(R \cup \set{b}) = \tr$ (Definition \ref{d:supattlinks}). As $C_a(R \cup \set{b}) = \tr$, from Definition \ref{d:xip}, we obtain there exists $B \in Sup_P(a)$ such that $R \cup \set{b} \subseteq \set{c \in \pr(a) \mid \neg c \not\in B}$. Then we can say there exists $B \in Sup_P(a)$ such that $R \subseteq \set{c \in \pr(a) \mid \neg c \not\in B}$. But then $C_a(R) = \tr$. An absurd! 
\end{proof}

\xipacceptance*

\begin{proof}

As $\Xi(P)$ is an $\nadf$, we obtain from Theorem \ref{t:nadf} that for every $a \in A$, 
\[ \varphi_a \equiv \bigvee_{R \in C^\mathit{max}_a}  \left( \bigwedge_{b \in \pr(a) - R} \neg b \right), \]
where $C^\mathit{max}_a = \set{R \in C^\tr_a \mid \textit{there is no } R' \in C^\tr_a \textit{ such that } R \subset R'}$. From Definition \ref{d:xip}, we know $C^\mathit{max}_a = \{ R \subseteq \set{b \in \pr(a) \mid \neg b \not\in B } \mid B \in \Sup_P(a) \textit{ and there is no } R' \in C^\tr_a$ \textit{such that} $R \subset R'\}$
$= \set{\set{b \in \pr(a) \mid \neg b \not\in B} \mid B \in \mathit{min}\set{\Sup_P(a)}}$, in which $\mathit{min}\set{\Sup_P(a)}$ returns the minimal sets (w.r.t. set inclusion) of $\Sup_P(a)$. Thus for every $a \in A$,
\[ \varphi_a \equiv \bigvee_{R \in C^\mathit{max}_a}  \left( \bigwedge_{b \in \pr(a) - R} \neg b \right) \equiv \bigvee_{B \in \mathit{min}\set{\Sup_P(a)}}  \left( \bigwedge_{\neg b \in B} \neg b \right), \]
But then, we obtain 
\[ \varphi_a \equiv \bigvee_{B \in \mathit{min}\set{\Sup_P(a)}}  \left( \bigwedge_{\neg b \in B} \neg b \right) \equiv \bigvee_{B \in \Sup_P(a)}  \left( \bigwedge_{\neg b \in B} \neg b \right). \]
\end{proof}

\pstable*

\begin{proof}


Let $P$ be an $\nlp$ and $\Xi(P) = (A, L, C^\tr)$ be the corresponding $\nadf$. Let $v$ be a 3-valued interpretation. We will prove $v$ is a partial stable model of $P$ iff $v$ is a complete model of $\Xi(P)$, i.e., $\Omega_P(v) = v$ iff for each $a \in A$, $v(a) = v(\varphi_a)$. 

We will prove by induction on $j$ that for each $a \in A$, $\Psi^{\uparrow\ j}_{\frac{P}{v}}(a) = \tr$ iff there exists $\Sup_P(r) \in \Sup_P(a)$ such that for each $x \in \Sup^{\uparrow\ j}_P(r)$, $v(x) = \tr$.

\begin{description}
\item [Base Case: ] We know  $\Psi^{\uparrow\ 1}_{\frac{P}{v}}(a) = \tr$ iff $a \in \frac{P}{v}$ iff there is a rule $a \leftarrow \naf b_1, \ldots, \naf b_n \in P$ ($n \geq 0$) such that for each $b_i$, ($1 \leq i \leq n$), $v(b_i) = \fa$ iff there exists $\Sup_P(r) \in \Sup_P(a)$ such that $\Sup^{\uparrow\ 1}_P(r) = \set{\neg b_1, \ldots, \neg b_n}$ and for each $\neg b_i \in \Sup^{\uparrow\ 1}_P(r)$, $v(\neg b_i) = \tr$.

\item [Inductive Hypothesis:] Assume for each $a' \in A$, $\Psi^{\uparrow\ n}_{\frac{P}{v}}(a') = \tr$ iff there exists $\Sup_P(r) \in \Sup_P(a')$ such that for each $x \in \Sup^{\uparrow\ n}_P(r)$, $v(x) = \tr$.

\item [Inductive Step:] We will prove $\Psi^{\uparrow\ n + 1}_{\frac{P}{v}}(a) = \tr$ iff there exists $\Sup_P(r) \in \Sup_P(a)$ such that for each $x \in \Sup^{\uparrow\ n + 1}_P(r)$, $v(x) = \tr$:

We know $\Psi^{\uparrow\ n + 1}_{\frac{P}{v}}(a) = \tr$ iff there exists $a \leftarrow a_1, \ldots, a_m \in \frac{P}{v}$ such that for each $a_i$, $1 \leq i \leq m$,  $\Psi^{\uparrow\ n}_{\frac{P}{v}}(a_i) = \tr$ iff there exists $a \leftarrow a_1, \ldots, a_m, \naf b_1, \ldots, \naf b_n \in P$ such that for each $a_i$, $1 \leq i \leq m$,  $\Psi^{\uparrow\ n}_{\frac{P}{v}}(a_i) = \tr$, and for each $b_j$, $1 \leq j \leq n$, $v(b_j) = \fa$ iff according to the Inductive Hypothesis,  there exists $a \leftarrow a_1, \ldots, a_m, \naf b_1, \ldots, \naf b_n \in P$ such that for each $a_i$, $1 \leq i \leq m$,
there exists $\Sup_P(r_i) \in \Sup_P(a_i)$ such that for each 
$x \in \Sup^{\uparrow\ n}_P(r_i)$, $v(x) = \tr$, and for each $b_j$, $1 \leq j \leq n$, $v(b_j) = \fa$ iff there exists $a \leftarrow a_1, \ldots, a_m, \naf b_1, \ldots, \naf b_n \in P$ and there are statements $r$, $r_i$, ($1 \leq i \leq m$) in $P$ with $\Conc_P(r) = a$ and $\Conc_P(r_i) = a_i$ such that for each $r_i$, for each $x \in \Sup^{\uparrow\ n}_P(r_i)$, $v(x) = \tr$, and for each $b_j$, $1 \leq j \leq n$, $v(\neg b_j) = \tr$ iff there exists $\Sup_P(r) \in  \Sup_P(a)$ such that for each $x \in \Sup^{\uparrow\ n + 1}_P(r)$, $v(x) = \tr$.
\end{description}

The above result guarantees for a 3-valued interpretation $v$ of $P$, $\Omega_P(v)(a) = \tr$ iff there exists $B = \Sup_P(r) \in \Sup_P(a)$ such that for each $x \in B$, $v(x) = \tr$, i.e., 

\begin{align}\label{eq:varphi-tr}
\Omega_P(v)(a) = \tr \textit{ iff } v\left( \bigvee_{B \in \Sup_P(a)} \left( \bigwedge_{\neg b \in B} \neg b  \right) \right) = \tr \textit{ iff }  v(\varphi_a) = \tr.
\end{align}

Similarly now we will prove by induction on $j$ that for each $a \in A$, $\Psi^{\uparrow\ j}_{\frac{P}{v}}(a) \neq \fa$ iff there exists $\Sup_P(r) \in \Sup_P(a)$ such that for each $x \in \Sup^{\uparrow\ j}_P(r)$, $v(x) \neq \fa$.

\begin{description}
\item [Base Case: ] We know  $\Psi^{\uparrow\ 1}_{\frac{P}{v}}(a) \neq \fa$ iff either $a \in \frac{P}{v}$ or $a \leftarrow \un \in \frac{P}{v}$ iff there exists a rule $a \leftarrow \naf b_1, \ldots, \naf b_n \in P$ ($n \geq 0$) such that for each $b_i$, ($1 \leq i \leq n$), $v(b_i) \neq \tr$ iff there exists $\Sup_P(r) \in  \Sup_P(a)$ such that $\Sup^{\uparrow\ 1}_P(r) = \set{\neg b_1, \ldots, \neg b_n}$ and for each $b_i$, ($1 \leq i \leq n$), $v(b_i) \neq \tr$ iff there exists $\Sup_P(r) \in \Sup_P(a)$ such that for each $\neg b_i \in \Sup^{\uparrow\ 1}_P(r)$, $v(\neg b_i) \neq \fa$.

\item [Inductive Hypothesis:] Assume for each $a' \in A$, $\Psi^{\uparrow\ n}_{\frac{P}{v}}(a') \neq \fa$ iff there exists $\Sup_P(r) \in \Sup_P(a')$ such that for each $x \in \Sup^{\uparrow\ n}_P(r)$, $v(x) \neq \fa$.

\item [Inductive Step:] We will prove $\Psi^{\uparrow\ n + 1}_{\frac{P}{v}}(a) \neq \fa$ iff there exists $\Sup_P(r) \in \Sup_P(a)$ such that for each $x \in \Sup^{\uparrow\ n + 1}_P(r)$, $v(x) \neq \fa$:

We know $\Psi^{\uparrow\ n + 1}_{\frac{P}{v}}(a) \neq \fa$ iff there exists $a \leftarrow a_1, \ldots, a_m \in \frac{P}{v}$ such that for each $a_i$, $1 \leq i \leq m$,  $\Psi^{\uparrow\ n}_{\frac{P}{v}}(a_i) \neq \fa$ iff there exists $a \leftarrow a_1, \ldots, a_m, \naf b_1, \ldots, \naf b_n \in P$ such that for each $a_i$, $1 \leq i \leq m$,  $\Psi^{\uparrow\ n}_{\frac{P}{v}}(a_i) \neq \fa$, and for each $b_j$, $1 \leq j \leq n$, $v(b_j) \neq \tr$ iff according to the Inductive Hypothesis, there exists $a \leftarrow a_1, \ldots, a_m, \naf b_1, \ldots, \naf b_n \in P$ such that for each $a_i$, $1 \leq i \leq m$,
there exists $\Sup_P(r_i) \in \Sup_P(a_i)$ such that for each 
$x \in \Sup^{\uparrow\ n}_P(r_i)$, $v(x) \neq \fa$, and for each $b_j$, $1 \leq j \leq n$, $v(b_j) \neq \tr$ iff there exists $a \leftarrow a_1, \ldots, a_m, \naf b_1, \ldots, \naf b_n \in P$ and there are statements $r$, $r_i$, ($1 \leq i \leq m$) in $P$ with $\Conc_P(r) = a$ and $\Conc_P(r_i) = a_i$ such that for each $r_i$, for each $x \in \Sup^{\uparrow\ n}_P(r_i)$, $v(x) \neq \fa$, and for each $b_j$, $1 \leq j \leq n$, $v(\neg b_j) \neq \fa$ iff there exists $\Sup_P(r) \in  \Sup_P(a)$ such that for each $x \in \Sup^{\uparrow\ n + 1}_P(r)$, $v(x) \neq \fa$.
\end{description}

The above result guarantees for a 3-valued interpretation $v$ of $P$, $\Omega_P(v)(a) \neq \fa$ iff there exists $B = \Sup_P(r) \in \Sup_P(a)$ such that for each $x \in B$, $v(x) \neq \fa$, i.e., 

\begin{align}\label{eq:varphi-fa}
\Omega_P(v)(a) = \fa \textit{ iff } v\left( \bigvee_{B \in \Sup_P(a)} \left( \bigwedge_{\neg b \in B} \neg b  \right) \right) = \fa \textit{ iff } v(\varphi_a) = \fa.
\end{align}

From (\ref{eq:varphi-tr}) and (\ref{eq:varphi-fa}), we conclude $v$ is a partial stable model of $P$ iff for all $a \in A$, $v(a) = \Omega_P(v)(a) = v\left( \bigvee_{B \in \Sup_P(a)} \left( \bigwedge_{\neg b \in B} \neg b  \right) \right) = v(\varphi_a)$, i.e., $v$ is a complete model of $\Xi(P)$.
\end{proof}

\equivalence*

\begin{proof}
This proof is a straightforward consequence from Theorem \ref{t:pstable}:

\begin{itemize}
    \item $v$ is a well-founded model of $P$ iff $v$ is the $\leq_i$-least partial stable model of $P$ iff (according to Theorem \ref{t:pstable}) $v$ is the $\leq_i$-least complete model of $\Xi(P)$ iff $v$ is the grounded model of $\Xi(P)$.
    \item $v$ is a regular model of $P$ iff $v$ is a $\leq_i$-maximal partial stable model of $P$ iff (according to Theorem \ref{t:pstable}) $v$ is a $\leq_i$-maximal complete model of $\Xi(P)$ iff $v$ is a preferred model of $\Xi(P)$.
    \item $v$ is a stable model of $P$ iff $v$ is a partial stable model of $P$ such that $\unk(v) = \set{s \in S \mid v(s) = \un} = \emptyset$ iff (according to Theorem \ref{t:pstable}) $v$ is a complete model of $\Xi(P)$ such that $\unk(v) = \emptyset$ iff (based on Theorem \ref{t:stable}) $v$ is a stable model of $\Xi(P)$.
    \item $v$ is an L-stable model of $P$ iff $v$ is a partial stable model of $P$ with minimal $\unk(v) = \set{s \in S \mid v(s) = \un}$ (w.r.t. set inclusion) among all partial stable models of $P$ iff (according to Theorem \ref{t:pstable}) $v$ a complete model of $\Xi(P)$ with minimal $\unk(v)$ among all complete models of $P$ iff $v$ is an $L$-stable model of $\Xi(P)$.
\end{itemize}

\end{proof}

\subsection{Propositions and Proofs from Section \ref{s:related}:}

\xipII*

\begin{proof}

Firstly, let $P$ be an $\nlp$ defined over a set $A$ of atoms, where each rule is like $a \leftarrow \naf b_1, \ldots, \naf b_n$. We know from Definitions \ref{d:substatement} and \ref{d:support} $\Sup_P(a) = \{ \set{\neg b_1, \ldots, \neg b_n} \mid a \leftarrow $ $\naf b_1, \ldots, \naf b_n \in P \}$. Then, according to Definition \ref{d:xip}, we obtain the $\adf$ $\Xi(P) = (A, L, C^\tr)$, where

\begin{itemize}
    \item $L = \set{(c,a) \mid a \leftarrow \naf b_1, \ldots, \naf b_n \in P \textit{ and } c \in \set{b_1, \ldots, b_n} }$;
    \item For $a \in A$, $C^\tr_a \hspace{-.1em} = \hspace{-.1em} \set{B' \subseteq \set{b \in \pr(a)\hspace{-.1em} \mid\hspace{-.1em} \neg b \not\in \set{b_1,\ldots,b_n} \hspace{-.1em} \mid \hspace{-.1em} a \leftarrow \naf b_1, \ldots, \naf b_n \in P } }$
    $= \set{B' \subseteq \pr(a) \mid a \leftarrow \naf b_1, \ldots, \naf b_n \in P \textit{ and } \set{b_1, \ldots, b_n } \cap B'= \emptyset }$.
\end{itemize}

According to Definition \ref{d:xip2}, we obtain the $\adf$ $\Xi_2(P) = (A, L_2, C^\tr_2)$, where

\begin{itemize}
\item $L_2  = \set{(c,a)  \mid  a \leftarrow \naf b_1, \ldots, \naf b_n  \in P \textit{ and } c \in \set{b_1, \ldots, b_n}} = L$;
\item For each $a \in A$, ${C^\tr_2}_a = \{B' \in \pr(a) \mid a \leftarrow \naf b_1, \ldots, \naf b_n \in P, \set{b_1, \ldots, b_n} \cap B' = \emptyset \} = C^\tr_a$.
\end{itemize}

Hence, $\Xi(P) = \Xi_2(P)$.
\end{proof}

\setaff*

\begin{proof}

In order to show $\mathit{DF}^{\mathit{SF}} = (A, L, C)$ is an $\nadf$, we will guarantee any $(r,s) \in L$ is an attacking link, i.e., for every $B \subseteq \pr(s)$, if $C_s(B \cup \set{r}) = \tr$, then $C_s(B) = \tr$:

Suppose $C_s(B \cup \set{r}) = \tr$. Then according to the translation from $\setaf$ to $\adf$, there is no $(X_i, s) \in R$ such that $X_i \subseteq B \cup \set{r}$. Thus there is no $(X_i, s) \in R$ such that $X_i \subseteq B$. This implies $C_s(B) = \tr$.
\end{proof}

\newpage



\end{document}